\pdfoutput=1
\documentclass[nohyperref]{article}

\usepackage[table]{xcolor}
\usepackage{microtype}
\usepackage{graphicx}
\usepackage{subfigure}
\usepackage{booktabs} 

\usepackage{hyperref}
\usepackage[utf8]{inputenc} 
\usepackage[T1]{fontenc}    
\usepackage{amsfonts}       

\usepackage[arxiv]{icml2022}

\usepackage{amsmath}
\usepackage{amssymb}
\usepackage{mathtools}
\usepackage{amsthm}
\usepackage{wrapfig}
\usepackage{soul}

\usepackage{tabularx}
\usepackage{multirow}
\usepackage{verbatim}
\usepackage{placeins}
\usepackage[font=small]{caption}
\usepackage[outercaption]{sidecap}
\usepackage{diagbox}
\usepackage{xspace}
\usepackage{adjustbox}
\usepackage{slashbox}
\usepackage[shortlabels]{enumitem}
\usepackage{mdwlist}
\usepackage{nicefrac}
\usepackage{bbm}
\usepackage{algorithm}
\usepackage{soul}
\usepackage{mathrsfs}
\usepackage{rotating}
\usepackage{bbm}

\usepackage[capitalize,noabbrev]{cleveref}

\theoremstyle{plain}
\newtheorem{theorem}{Theorem}[section]

\theoremstyle{definition}

\theoremstyle{remark}

\usepackage[textsize=tiny]{todonotes}

\usepackage{amsmath,amsfonts,bm}

\def\eqref#1{equation~\ref{#1}}

\def\1{\bm{1}}

\def\rvb{{\mathbf{b}}}

\def\rvf{{\mathbf{f}}}

\def\rvw{{\mathbf{w}}}
\def\rvx{{\mathbf{x}}}

\def\rvmu{{\boldsymbol{\mu}}}

\def\rmA{{\mathbf{A}}}

\def\rmD{{\mathbf{D}}}

\def\rmF{{\mathbf{F}}}

\def\rmI{{\mathbf{I}}}

\def\rmS{{\mathbf{S}}}

\def\rmW{{\mathbf{W}}}

\DeclareMathAlphabet{\mathsfit}{\encodingdefault}{\sfdefault}{m}{sl}
\SetMathAlphabet{\mathsfit}{bold}{\encodingdefault}{\sfdefault}{bx}{n}

\def\gE{{\mathcal{E}}}

\def\gG{{\mathcal{G}}}

\def\gL{{\mathcal{L}}}

\def\gN{{\mathcal{N}}}

\def\gV{{\mathcal{V}}}

\def\sP{{\mathbb{P}}}

\def\sR{{\mathbb{R}}}

\def\s1{{\mathbbm{1}}}

\newcommand{\E}{\mathbb{E}}

\newcommand{\Tau}{\pmb{\mathcal{T}}}

\newcommand{\method}{GGCN\xspace}

\newcommand{\ratioi}{\overline{r_i}}
\newcommand{\heterophilyil}{\widehat{h}_i^l}

\icmltitlerunning{Heterophily and Oversmoothing in Graph Convolutional Neural Networks}

\begin{document}

\twocolumn[
\icmltitle{Two Sides of the Same Coin: \\ Heterophily and Oversmoothing in Graph Convolutional Neural Networks}



\icmlsetsymbol{equal}{*}

\begin{icmlauthorlist}
\icmlauthor{Yujun Yan}{um}
\icmlauthor{Milad Hashemi}{g}
\icmlauthor{Kevin Swersky}{g}
\icmlauthor{Yaoqing Yang}{ucb}
\icmlauthor{Danai Koutra}{um}
\end{icmlauthorlist}

\icmlaffiliation{um}{University of Michigan, Ann Arbor}
\icmlaffiliation{g}{Google Research}
\icmlaffiliation{ucb}{University of California, Berkeley}

\icmlcorrespondingauthor{Yujun Yan}{yujunyan@umich.edu}


\icmlkeywords{Graph Neural Networks, Heterophily, Oversmoothing}

\vskip 0.3in
]



\printAffiliationsAndNotice{}

\begin{abstract}

In node classification tasks, graph convolutional neural networks (GCNs) have demonstrated competitive performance over traditional methods on diverse graph data.  However, it is known that the performance of GCNs degrades with increasing number of layers (oversmoothing problem) and recent studies have also shown that GCNs may perform worse in heterophilous graphs, where neighboring nodes tend to belong to different classes (heterophily problem). These two problems are usually viewed as unrelated, and thus are studied independently, often at the graph filter level from a spectral perspective.

{We are the first to take a unified perspective to jointly explain the oversmoothing and heterophily problems at the node level.}  
Specifically, we profile the nodes via two quantitative metrics: the relative degree of a node (compared to its neighbors) and the node-level heterophily. 
Our theory shows that the interplay of these two profiling metrics defines three cases of node behaviors, 
which explain the oversmoothing and heterophily problems jointly and can predict the performance of GCNs. 
Based on insights from our theory, we show theoretically and empirically the effectiveness of two strategies: structure-based edge correction, which learns corrected edge weights from structural properties (i.e., degrees), and feature-based edge correction, which learns signed edge weights from node features.
Compared to other approaches, which tend to handle well either heterophily or oversmoothing, we show that {our model, GGCN}, which incorporates the two strategies performs well in both problems. Codes are available at \href{https://github.com/Yujun-Yan/Heterophily\_and\_oversmoothing}{this link}.

\end{abstract}
\section{Introduction}
\label{section: introduction}

Recently, GCNs~\cite{defferrard2016convolutional, kipf2016semi, velivckovic2017graph} have been widely used in 
applications ranging from social science~\cite{li2019encoding} and biology~\cite{yan2019groupinn} to program understanding~\cite{allamanis2018learning,shi2019learning}. A typical GCN architecture~\cite{gilmer2017neural} for the node classification task can be decomposed into two {main components}: propagation/aggregation, and combination. 
Messages are first exchanged between neighboring nodes, then aggregated. 
Afterwards, they are combined with the self-representations (a.k.a., the current node representations) to update the node representations. 
{Though GCNs are generally effective, their performance may degrade in some cases.}

{\citeauthor{li2018deeper}~\cite{li2018deeper} found that GCNs perform worse with increasing number of layers, which is termed as the ``oversmoothing problem''.}
Recent works claim that oversmoothing could be caused by GCNs exponentially losing expressive power in the node classification task~\cite{oono2019graph} and that the node representations converge to a stationary state which is decided by the degree of the nodes and the input features~\cite{chen2020simple, wang2019improving, rong2019dropedge,RossiJSAKL2020}. 
{These works analyze the asymptotic node representations in the limit of infinite layers, but they do not characterize how the node representations change over the layers ({we call different types of changes \textit{node behaviors}}) and how different node behaviors contribute to the oversmoothing problem.
\citeauthor{chen2020measuring}~\cite{chen2020measuring} empirically define metrics to measure oversmoothing, {but it remains unclear what causes it and how the metrics are related in theory.}} 
{Going beyond empirical definitions,} we propose theoretically-grounded node-level metrics that characterize different node behaviors across GCN layers, and show theoretically and empirically how they can explain the oversmoothing problem and identify the nodes that trigger it. 
{GCNs may also perform poorly}
on heterophilous graphs~\cite{pei2019geom, lim2021large}, 
which---unlike homophilous graphs---comprise 
many neighboring nodes that belong to different classes~\cite{newman2002assortative}. {This is termed as the ``heterophily problem''.}
For instance,
in protein networks, amino acids of different types tend to form links~\cite{zhu2020beyond}, and in transaction networks, fraudsters are more likely to connect to accomplices than to other fraudsters~\cite{pandit2007netprobe}. Most GCNs~\cite{kipf2016semi, velivckovic2017graph} fail to effectively capture heterophily, 
{so various designs have been proposed to handle it~\cite{pei2019geom, zhu2020beyond, chien2021adaptive, bo2021beyond}. These works  take the spectral perspective and design various high-frequency graph filters to address heterophily. However, they neglect the fact that different node behaviors impact  GCNs' performance under heterophily differently {and  need to be handled differently}. In this work, we show that GCNs can perform differently on graphs that have similar {graph-level heterophily} but are dominated by different node behaviors.}

{These two problems, which cause performance degradation, }have mostly been studied independently. Recent work on oversmoothing~\cite{chen2020simple} was shown only empirically to address heterophily, and vice versa~\cite{chien2021adaptive}. 
Motivated by this empirical observation, 
we are the first to {find a joint explanation for the two problems. Specifically, we aim to identify meaningful {node-level} metrics that are theoretically-grounded and their interplay can be used to characterize different node behaviors {(profiles)}, which in turn can explain both problems.
We found that the relative degree of a node (compared to its neighbors) and its node-level heterophily define three types of node behaviors, two of which are related to performance degeneration. 
{Based
on our theoretical insights, we show theoretically
and empirically the effectiveness of two strategies: 
{structure-based edge correction, which learns corrected edge weights from structural properties like degrees, and feature-based edge correction, which learns signed edge weights.}
{Signed edge weights can model both positive and negative influence from the neighbors. Though prior work ~\cite{chien2021adaptive,bo2021beyond} suggests using negative coefficients for graph filters to capture the "negative influence" (termed as "signed messages"), we introduce a new, {more powerful} design, which is derived from our node-level analysis instead of the typically-used spectral analysis. }}}

In sum, we make the following contributions:
\setlist{leftmargin=*}

\begin{itemize*}
    \vspace{-0.25cm}
    \item \textbf{Theoretically-grounded Node Metrics:}{ We introduce two theoretically-grounded metrics, relative degree and node-level
heterophily, to profile the nodes across layers in GCNs. The profiling provides a joint explanation for what triggers the heterophily and oversmoothing problems.
    }
    \item \textbf{Insights:} 
    Our theory states that {under certain conditions,
    low-degree nodes tend to trigger the oversmoothing problem in strongly homophilous graphs}, while high-degree nodes tend to {cause the oversmoothing and heterophily problems in weakly homophilous (i.e., heterophilous) graphs.}
    {We also show that using signed edge weights can be helpful in alleviating both problems.}
    \item \textbf{{Improved  Model \& Empirical Analysis:}}  
    Based on our insights, we show theoretically and empirically the effectiveness of two strategies: 
    {structure-based edge correction, which learns edge weights from structural properties, and feature-based edge correction, which learns signed edge weights from node features.}
    {Our empirical results show that our model, \method, which leverages the two strategies is robust to oversmoothing, achieves state-of-the-art performance on datasets with high levels of heterophily, and achieves competitive performance on homophilous datasets.} 
\end{itemize*}

\section{Preliminaries}
\label{section: background}
We first provide the notations \& definitions that we use in the paper, and a brief background on GCNs.

\textbf{Notation.}
We denote an unweighted and self-loop-free graph as $\gG$ ($\gV$, $\gE$) and its adjacency matrix as $\rmA$. 
We represent the degree of node $v_i\in\gV$ by $d_i$, and the degree matrix---which is a diagonal matrix whose elements are node degrees---by $\rmD$. 
Let $\gN_i$ be the set of nodes directly connected to $v_i$, i.e., its neighbors. $\rmI$ is the identity matrix.
{We denote the node representations at $l$-th layer as $\rmF^{(l)}$, and the $i$-th row of $\rmF$ is $\rvf_i^{(l)}$, which is the representation of node $v_i$. The input node features are given by $\rmF^{(0)}$. The weight matrix and bias vector at the $l$-th layer are denoted as $\rmW^{(l)}$ and $\rvb^{(l)}$, respectively.} 


\textbf{Supervised Node Classification Task.} We focus on node classification: 
{Given a random sample of node representations $\{\rvf_1^{(0)}\dots\rvf_n^{(0)}\} \in \sR^m$ and their labels $\{y_1\dots y_n\} \in \sR^n$ for training, we aim to learn a function $\mathscr{F}:\sR^m \mapsto \sR^n$,  such that the loss $\E(\mathscr{L}(y_i, \hat{y_i}))$ is minimized, where $\hat{y_i}=\mathscr{F}(\rvf_i^{(0)})$ is the predicted label of $v_i$. The \textbf{misclassification rate} is defined as the probability $\sP(y_i \neq \hat{y_i})$ to misclassify an arbitrary node in the node representation space.} 


\textbf{GCNs.}
In node classification tasks, an $L$-layer GCN contains two components \citep{gilmer2017neural}: (1) neighborhood propagation and aggregation: $\widehat{\rvf_i^{(l)}}$ = \texttt{AGGREGATE}($\rvf_j^{(l)}$, $v_j \in \gN_i$), and (2) combination: $\rvf_i^{(l+1)}$ = \texttt{COMBINE}($\widehat{\rvf_i^{(l)}}$, $\rvf_i^{(l)}$), where {\texttt{AGGREGATE} and \texttt{COMBINE} are learnable functions. The loss is given by $\mathscr{L}_{CE}$=\texttt{CrossEntropy}(\texttt{Softmax}($\rvf_i^{(L)}\rmW^{(L)}+\rvb^{(L)}$), $y_i$). }
The vanilla GCN suggests a renormalization trick on the adjacency $\rmA$ to prevent gradient explosion~\citep{kipf2016semi}. The $(l+1)$-th output is given by: $\rmF^{(l+1)} = \sigma(\tilde{\rmA}\rmF^{(l)}\rmW^{(l)})$, where $\tilde{\rmA} = \tilde{\rmD}^{-1/2}(\rmI+\rmA)\tilde{\rmD}^{-1/2}$, $\tilde{\rmD}$ is the degree matrix of $\rmI+\rmA$, and $\sigma$ is ReLU. {When the non-linearities in the vanilla GCN are removed, it reduces to a linear model called SGC~\citep{wu2019simplifying}, which has competitive performance and is widely used in theoretical analyses~\citep{oono2019graph, chen2020simple}. For SGC, the $l$-th layer representations are given by: $\rmF^{(l)}=\tilde{\rmA}^{l}\rmF^{(0)}$ and the last layer is a logistic-regression layer: $\hat{y_i} = \texttt{Softmax}(\rmF^{(L)}\rmW^{(L)}+\rvb^{(L)}).$ {We note that only one weight matrix $\rmW^{(L)}$ is learned, which  is equivalent to the products of all weight matrices in a linear GCN.} 
More related works can be found in~\S~\ref{section:related_work}.} 

\begin{figure*}[t]
\centering
\includegraphics[scale=0.21, bb=1250 0 300 250]{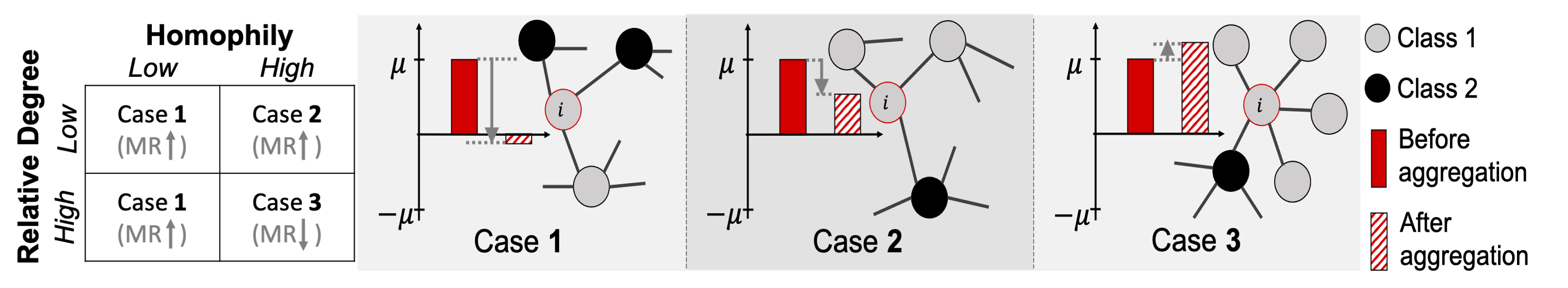}
\vspace{-0.2cm}
\caption{Node representation dynamics during neighborhood aggregation (in 1D for illustration purposes; `MR': misclassification rate). 
The expectation of node representations from class 1 \& 2 are denoted by $\mu$ and $-\mu$, respectively.
The bars show the expected node representations of node $v_i$ before and after the aggregation.} 
\label{fig:concept}
\vspace{-0.6cm}
\end{figure*}
\section{Theoretical Analysis}
\label{section: analysis}

{In this section, we formally introduce two metrics: node-level homophily $h_i$ and relative degree $\ratioi$. We show theoretically 
(1)~how the two metrics and their extensions (effective homophily $\heterophilyil$ and effective relative degree $\bar{r_i}^{l}$) characterize different node behaviors across GCN layers, and 
(2)~how this node profiling can be used to explain the oversmoothing and heterophily problems. }

{To begin with, we first introduce the theoretical setup. Throughout the section, we analyze binary node classification using the {typically-studied} SGC model (\S~\ref{section: background}). The nodes in class 1 are denoted as set $\gV_{1}$ and nodes in class 2 are denoted as set $\gV_{2}$. 
Later, in \S~\ref{section: experiments}, we show empirically that the insights obtained in this section are effective for other non-linear models in multi-class classification.}
\subsection{Assumptions} 
{\textbf{Notations.} We use ``i.d.'' 
to represent random variables / vectors that follow the same marginal distribution and their joint probability density function (PDF) is a permutation-invariant function
$p(\rvx_1, \dots, \rvx_n)=p(\textbf{P}(\rvx_1, \dots, \rvx_n))$, where $\textbf{P}$($\cdot$) means permutation. We use {$\E_{A|B}{(\cdot)}$ to denote the expectation  taken over the randomness of $A$ given $B$.} 
}

We make the following assumptions:

\noindent \textbf{(1) Random Graph}: Node degrees $\{d_i\}$ are i.d.\ random variables, where $\{(\cdot)_i\}$ represents a set with $i=1, \dots, |\gV|$.


\noindent {\textbf{(2) Inputs}: (2.1) Node labels $\{y_i\}$ are i.d.\ Bernoulli random variables given by the ratio $\rho$: $\rho \equiv \frac{\sP(y_i=1)}{\sP(y_i=2)}, \forall i.$ The event $\{y_i=y_j|v_j\in\gN_i\}$ is independent of $y_i$, $\forall i, j$. \\
(2.2) Initial input node features $\{\rvf_i^{(0)}\}$ are random vectors  given by (PDF)  $f({\rvx})$, which is expressed as: \\ 
\begin{center}
\vspace{-0.6cm}
$f({\rvx})=\begin{cases}
f_1(\rvx),  \text{when } y_i=1.\\
f_2(\rvx),  \text{when } y_i=2.
\end{cases}$ 
\vspace{-0.3cm}
\end{center}

$\E(\rvf_i^{(0)}|y_i)=\begin{cases} \rvmu, & \text{when } y_i=1\\ -\rho\rvmu, & \text{when } y_i=2  
\end{cases},$ so $\E(\rvf_i^{(0)})$=$\mathbf{0}$. 
}

\noindent \textbf{(3) Independence}: $\rmA$ is independent of $\{y_i\}$ and $\{\rvf_i^{(0)}\}$. $\forall i, j$ given $y_i$, $\rvf_i^{(0)}$ and $y_j$ are conditional independent.
\subsection{Node-level Metrics: Definitions}
\label{sec:metrics}
\textbf{Node-level Homophily and Heterophily.} Given a set of node labels/classes, \textit{homophily} captures the tendency of a node to have the same class as its neighbors. 
Specifically, the homophily of node $v_i$ is defined as:  
\vspace{-0.2cm}
$$h_i \equiv \sP(y_i=y_j|v_j \in \gN_i).\vspace{-0.2cm}$$
High homophily corresponds to low heterophily, and vice versa, so we use these terms interchangeably. 

\textbf{Relative Degree $\ratioi$.}
{The relative degree of node $v_i$ is:
\vspace{-0.35cm}
$$\ratioi \equiv \E_{\rmA|d_i}(\frac{1}{d_i} \sum_{j \in \gN_i} r_{ij}|d_i), \text{where } r_{ij}\equiv\sqrt{\frac{d_i + 1}{d_j + 1}}.\vspace{-0.35cm}$$ It evaluates the node degree compared to its neighbors' degrees. When all the nodes have the same degree, $\ratioi=1.$}
\subsection{Node Profiling}
\label{subsec:node_profile}
{In this section, we theoretically show how the two metrics can characterize different node behaviors across layers.}\\
{\textbf{Movements of Node Representations.} We monitor the node behaviors by tracking the changes of node representations across the layers. 
Each node representation is mapped to a point in the feature space whose coordinates are decided by the representation vector. In this way, the changes of a node's representations across the layers can be viewed as the movements of the mapped point. For example, $\rvf_i^{(l+1)}-\rvf_i^{(l)}$ is referred to as the movement of node $v_i$'s representation at the $l$-th layer. Next, we show that the interplay of the two metrics relates to different types of movements.}

\subsubsection{Movements at the initial layer}
\label{sec:initial-stage}
{We examine how the node representations change in expectation. Without loss of generality, we assume $v_i \in \gV_1$, the other case can be derived similarly.}
\begin{theorem}
Given $v_i \in \gV_1$ and $d_i$, the conditional expectation of representation $\rvf_i^{(1)}$ is given by:
\vspace{-0.2cm}
\begin{align}
&\E_{\rmA, \{y_i\}, \{\rvf_i^{(0)}\}|d_i, v_i \in \gV_1}(\rvf_i^{(1)}|v_i \in \gV_1, d_i) \nonumber \\&=  
  \left (\frac{((1+\rho)h_i-\rho) d_i \ratioi + 1}{d_i + 1}\right )\E(\rvf^{(0)}_i|v_i \in \gV_1) \nonumber\\&\equiv \gamma_i^1 \E(\rvf^{(0)}_i|v_i \in \gV_1),
  \label{eq:factor}
\end{align}
where the multiplicative factor $\gamma_i^1$ is: 
\vspace{-0.2cm}
\begin{equation}
\gamma_i^1 \in 
\begin{cases}
  (-\infty, \frac{1}{2}], & \text{if}\ h_i\leq \frac{\rho}{1+\rho} \\
  {(0, 1]}, & \text{if}\ h_i>\frac{\rho}{1+\rho} \ \mathrm{\&}\  \ratioi \leq \frac{1}{(1+\rho)h_i-\rho}\\
  (1, \infty), & \mathrm{otherwise}.
\end{cases}
\label{eq:initial_stage}
\vspace{-0.3cm}
\end{equation}
In case 1, $\gamma_i^1$ decreases as $d_i$ increases; In case 3, $\gamma_i^1$ increases as $d_i$ increases.
\label{theorem: initial_stage}
\end{theorem}
\vspace{-0.5cm}
\begin{proof}
We provide the proof in App.~\ref{app:proof1}.
\end{proof}
\vspace{-0.5cm}
{From Thm.~\ref{theorem: initial_stage}, we identify three types of movements of node representations, which are characterized by relative degree $\ratioi$ and homophily level $h_i$. For illustration purposes, in Fig.~\ref{fig:concept}, we illustrate the three cases when {we apply our theorem to 1D node representations.}
The bars reflect the value change of $v_i$'s node representation.} 
{Intuitively, under heterophily (case 1), node representations tend to move closer to the representations of the other class. The higher the degree, the more the representation moves. Under high homophily but low degrees (case 2), node representations still tend to move towards the other class, but not as much as in case 1. Only when both the homophily and the degree is high, the node representations may move away from the other class. Thus, case 3 is the only favorable case.} 
\subsubsection{Movements at deeper layers}
\label{sec:developing-stage}
{The scenarios at deeper layers are more complex. However, by extending the definitions of the two metrics, we can obtain a similar equation, and the extended metrics can characterize the nodes into 3 cases similar to Thm.~\ref{theorem: initial_stage}.}

Based on Thm.~\ref{theorem: initial_stage}, message passing scales the node representations ($\gamma_i^1$).
To account for the accumulated scaling effects, 
let $\xi^l_i$ be a discount factor at the $l$-th layer and let:
 $\mathbb{E}_{\rmA,\{y_i\},\{\rvf_i^{(0)}\}|d_i, y_i}\left ({\mathbf{f}_i^{(l)}}|d_i, y_i \right)=\xi_{i}^l\E(\rvf^{(0)}_i|y_i).$
Conditioned on $d_i$, $\rmA$ and $y_i$, the nodes contributing positively are defined as  
{\small $\hat{\mathcal{N}_i^s}(\rmA, y_i, d_i) \equiv 
\{v_j|v_j \in \gN_i \text{ and } \rvf_i^{(l)}\cdot{\rvf_j^{(l)}}^T>0\}$}.
We denote {\small $\mathbb{E}_{\rmA,\{y_i\},\{\rvf_i^{(0)}\}|d_i,y_i, v_j \in \gN_i}\Big(\mathbf{f}_j^{(l)}r_{ij}|d_i, y_i, v_j \in \gN_i\Big)=\begin{cases}
{\xi_{i}^{l}}'\E(\rvf^{(0)}_i|y_i) & v_j \in   \hat{\mathcal{N}_i^s}(\rmA, y_i, d_i) \\
-\rho_{i}^{l}{\xi_{i}^{l}}'\E(\rvf^{(0)}_i|y_i) & v_j \not\in   \hat{\mathcal{N}_i^s}(\rmA, y_i, d_i)
\end{cases}$}, where {${\xi_{i}^{l}}'$ represents the accumulated scaling effect when $v_j$ contributes positively to $v_i$. Due to our assumptions, neighbors are statistically indistinguishable, so ${\xi_{i}^{l}}'$ characterizes the neighborhood property of $v_i$; ${\rho_{i}^{l}}$ is the ratio of the two conditional expectations {when a neighbor contributes positively and negatively}.}
\vspace{0.15cm}

{In order to simplify the derivation and utilize a similar reasoning as in the initial layer, }
we extend the metrics in \S~\ref{sec:metrics}:
(1)~The \textbf{effective homophily} 
of node $i$ is defined as  {$\hat{h_i^{l}}=\sP(v_j \in \hat{\mathcal{N}_i^s}|v_j \in \gN_i, d_i, \xi_{i}^l)$---{i.e.,  the probability of a neighbor contributing positively. At the initial layer, it reduces to the node-level homophily $h_i$.}
(2)~The \textbf{effective relative degree}, {\small {$\bar{r_i}^{l}\equiv \frac{{\xi_{i}^{l}}'}{{\xi_{i}^{l}}}=\frac{\mathbb{E}_{\rmA,\{y_i\},\{\rvf_i^{(0)}\}|d_i,y_i, v_j \in \hat{\mathcal{N}_i^s}}\Big(\mathbf{f}_j^{(l)}r_{ij}|d_i, y_i, v_j \in \hat{\mathcal{N}_i^s}\Big)}{\mathbb{E}_{\rmA,\{y_i\},\{\rvf_i^{(0)}\}|d_i, y_i}\left ({\mathbf{f}_i^{(l)}}|d_i, y_i \right)}$}}, represents the ratio of expected $\rvf_j^{l}r_{ij}$ to expected $\rvf_i^{l}$, when node $v_j$ contributes positively to $v_i$. At the initial layer, it reduces to the node's relative degree, $\bar{r_i}$.
}


\begin{theorem}
Given $y_i$ and $d_i$, the conditional expectation of $\rvf_i^{(l+1)}$ is: 
\vspace{-0.6cm}
\begin{align}
&\mathbb{E}_{\rmA,\{y_i\},\{\rvf_i^{(0)}\}|d_i,y_i}\Bigg(\mathbf{f}_i^{(l+1)}|d_i,y_i)\Bigg)  \\
&= \frac{\left((\hat{h_i^{l}}(1+\rho_{i}^l)-\rho_{i}^l)d_i\bar{r_i}^{l}+1\right)}{d_i+1}\xi_{i}^l\E(\rvf_i^{(0)}|y_i) \\&\equiv \gamma_i^{(l+1)}\E(\rvf_i^{(0)}|y_i). \nonumber
\label{eq: developing stage}
\end{align}
\label{theorem: developing stage}
\vspace{-0.9cm}
\end{theorem}
\begin{proof}
\vspace{-0.2cm}
We provide the proof in App.~\ref{app:proof2}. 
\vspace{-0.1cm}
\end{proof}
\vspace{-0.1cm}

\vspace{-0.2cm}
{\subsubsection{Movements \& Misclassification Rate} In App.~\ref{app:mr}, we prove that under certain condition, the movement of representations towards the other class by a non-zero step increases the misclassification rate, causing performance degradation. {We note that the condition is important to explain why recent works~\cite{luan2021heterophily, ma2021homophily} find that GCNs can sometimes perform well in heterophilous graphs (e.g., bipartite graphs) because the representations of opposite classes swap places.}}

\subsubsection{Explanation for Heterophily and Oversmoothing}

 \noindent \underline{Oversmoothing problem:} {Nodes with low homophily (case 1) and nodes with high homophily but low degrees (case 2) cannot benefit from message aggregation. Their representations tend to move towards the other class. Under certain conditions, their misclassification rate is increased via message aggregation. GCNs' performance on node classification degrades each time the message aggregation is applied, which explains oversmoothing  in homophilous and heterophilous graphs. }

\noindent {\underline{Heterophily problem:} In heterophilous graphs, nodes from case 1 (and sometimes case 2) dominate. The performance degradation occurs at the first layer, which explains why GCNs may perform worse than MLP in heterophilous graphs.}

\noindent {\underline{Relation between the problems:} (1) In heterophilous graphs, both problems are caused by nodes from case 1 and case 2. Message aggregation makes the representations of these nodes (esp.\ case 1) less distinguishable. 
(2) In homophilous graphs, we can decompose the oversmoothing process into two stages, where the node behaviors in the second stage resemble those in the heterophily problem. \textit{Initial Stage.} At shallow layers, nodes of case 3 dominate initially, GCNs benefit from graph convolution. \textit{Developing Stage.} Nodes of case 2 and case 1 cannot benefit from message aggregation and their misclassification rate increases with more layers. In deep layers, they are misclassified and are wrongly viewed by their neighbors as coming from a different class. Thus, nodes of case 2 and case 1 cause low effective homophily
of their neighbors. At last, most nodes have low effective homophily in deep layers and are transformed to case 1, which resembles the phenomenon in heterophilous graphs ("pseudo-heterophily").
}
\subsection{Node Profiling With Signed Edges}
\label{subsec: sign_theory}
{In this section, we discuss how the interplay of the two metrics changes when {allowing} signed edges. 
We provide theory to show {when} 
signed {edges} can help enhance the performance in heterophilous graphs and  alleviate oversmoothing.
Due to limited space,
we only show the effect of signed {edges} at the initial layer; similar results can be derived in deeper layers.} 

\textbf{Setup.} 
{Each edge is assigned  a positive or a negative sign. Messages passing through a signed edge will be multiplied by its sign. Ideally, we would like to assign the positive signs to homophilous edges (i.e, edges connecting nodes from the same class) and negative signs to heterophilous edges.}
{In reality, we cannot access to the nodes' ground-truth labels and cannot know whether the edges are homophilous or heterophilous. Thus we learn the signs, which introduces errors.}
For node $v_i$, we define $m_i^l$ as the \textit{ratio of neighbors that send incorrect messages} at the $l$-th layer {because we wrongly assign a negative (positive) sign to a homophilous (heterophilous) edge that connects them.} 
We define the $l$-th layer \textit{error rate} as $e_i^l=\E(m_i^l)$, where the expectation is over the randomness of the neighbors that send incorrect messages. We assume that $m_i^l$ is independent of $\{d_i\}$, $\{y_i\}$ and $\{\rvf_i^{0}\}$.
\begin{theorem}
\textit{\textbf{[Signed Edges]}} By allowing {signed edges}, the {movements of representations} will {\textbf{be less affected} by the initial homophily level $h_i$, and will be dependent on the \textbf{error rate} $e_i^0$}. The multiplicative factor $\gamma_i^1$ at the first layer is given by:
\vspace{-0.1cm}
\begin{align}
&\E_{\rmA, \{y_i\}, \{\rvf_i^{(0)}\}|d_i, v_i \in \gV_1}(\rvf^{(1)}_i|d_i, v_i \in \gV_1) \\=    
\Bigg ( & \frac{(1 - 2e_i^0)(\rho+(1 -\rho)h_i)d_i\ratioi + 1}{d_i+1}\Bigg )\E(\rvf^{(0)}_i|v_i \in \gV_1), 
\nonumber \\
\text{where } \nonumber
\label{eq:signed_messages}
\vspace{-0.9cm}
\end{align}
\vspace{-0.9cm}
\begin{equation}
\vspace{-0.5cm}
\\ \gamma_i^1 \in 
\begin{cases}
  (-\infty, \frac{1}{2}], & \text{if}\ e_i^0 \geq 0.5 \\
  {(0, 1]}, & \text{if}\ e_i^0 < 0.5 \ \mathrm{\&}\  \ratioi \leq \frac{1}{(1 - 2e_i^0)(\rho+(1-\rho)h_i)}\\
  (1, \infty), & \mathrm{otherwise}.
\end{cases}
\label{eq: signed_message_initial_stage}
\end{equation}
\label{theorem: signed_messages}
\end{theorem}
\vspace{-0.6cm}
\begin{proof}
The proof is provided in App.~\ref{app:proof3}.
\end{proof}

\vspace{-0.4cm}
{
From Eq.~\ref{eq: signed_message_initial_stage}, we see that when using signed edges, to benefit from case 3 ($\gamma_i^1>1$), the minimum relative degree satisfies: $\ratioi > \frac{1}{(1 - 2e_i^0)(\rho+(1-\rho)h_i)}$. Given $h_i \leq 1$, if the error rate is low ($e_i^0 \ll 0.5$),  we get: $\frac{1}{(1 - 2e_i^0)(\rho+(1-\rho)h_i)} \leq \frac{1}{(1+\rho)h_i-\rho}$, and $\frac{1}{(1+\rho)h_i-\rho}$ is the minimum relative degree required when not using signed edges.
This implies that more nodes can benefit from using signed edges. 
\ul{We note that if low error rate cannot be guaranteed, signed edges may hurt the performance.}}
\section{Model Design}
\label{section: model_design}



Based on our theoretical analysis, we propose two new, simple mechanisms to address both the heterophily and oversmoothing problems: {structure-based edge correction} and {feature-based edge correction}. We integrate these mechanisms, along with a decaying combination of the current and previous node representations~\cite{chen2020simple},  into a generalized GCN model, \method, whose effectiveness we show empirically in \S~\ref{section: experiments}.


\subsection{Structure-based Edge Correction}
\label{subsec: degree}
Our analysis in \S~\ref{subsec:node_profile} and \S~\ref{subsec: sign_theory} highlights that, {when the homophily level is high (or error rate is low)}, oversmoothing is initially triggered by low-degree nodes. Thus, we aim to compensate for low degrees {by learning new edge weights. Unlike attention which encodes similarity of features, these weights only contain structural information (i.e., degrees).}

Based on Eq.~(\ref{eq: signed_message_initial_stage}), 
we require that the node degrees satisfy $\ratioi > \frac{1}{(1 - 2e_i^0)(\rho+(1-\rho)h_i)}$ to prevent oversmoothing.
Since the node degrees cannot be modified, our strategy is to \textit{rescale} or correct the {edge weights}  
by multiplying with scalars $\tau_{ij}^l$: 

\vspace{-0.25cm}
{\small
\begin{align}
     &(\tilde{A}{\rmF^{(l)}})[i,:]=\frac{{\rmF^{(l)}}[i,:]}{d_i+1} +  \boxed{\sum_{v_j \in \gN_i}{\frac{{\rmF^{(l)}}[j,:]}{\sqrt{d_i+1}\sqrt{d_j+1}}}} \nonumber \\ 
     & \Longrightarrow \sum_{v_j \in \gN_i}{\frac{\colorbox{gray!20}{$\tau_{i,j}^l$} {\rmF^{(l)}}[j,:]}{\sqrt{d_i+1}\sqrt{d_j+1}}}.
\end{align}
}
\vspace{-0.3cm}



This multiplication 
is equivalent to changing the ratio $r_{ij}$ in Thm.~\ref{theorem: initial_stage} to $\sqrt{\frac{(\tau_{ij}^l)^2(d_i + 1)}{d_j + 1}}$. 
That is, a larger $\tau_{ij}^l$ increases the effective $\ratioi$ at layer $l$.
Training independent $\tau_{i,j}^l$ is not practical because it would require $O(\lvert \mathcal{V}\rvert^2)$ additional parameters per layer, which can  lead to overfitting. Moreover, low-rank parameterizations suffer from unstable training dynamics. {Intuitively, when $r_{ij}$ is small, we would like to compensate for it via a larger $\tau_{i,j}^l$. Thus,} we set $\tau_{i,j}^l$ to be a function of $r_{ij}$ as follows:

\vspace{-0.45cm}
{\small
\begin{equation}
    \tau_{ij}^l = \text{softplus}\left (\lambda_0^l\left (\tfrac{1}{r_{ij}}-1\right )+\lambda_1^l\right ),
\vspace{-0.1cm}
\end{equation}
}
where $\lambda_0^l$ and $\lambda_1^l$ are learnable parameters. We subtract 1 so that when $r_{ij}=1$ (i.e., $d_i = d_j$), then $\tau_{ij}^l=\text{softplus}(\lambda_1^l)$ is a constant bias. 

{Let $\Tau^{(l)}$ be a matrix with elements $\tau_{ij}^l$. Our model GGCN learns a corrected adjacency matrix at $l$-th layer: 
\vspace{-0.25cm}
$$\small {\widehat{\tilde{\rmA}^{l}}} = \colorbox{gray!20}{$\tilde{\rmA} \odot \Tau^{(l)}$}$$
where $\odot$ is element-wise multiplication.}

\subsection{Feature-based Edge Correction}
\label{subsec: sign}
Theorem~\ref{theorem: signed_messages} points out the 
importance of signed {edges} in tackling the heterophily and oversmoothing problems.  Inspired by this, we {aim to learn the signed edge weights based on node features. Unlike attention weights that are usually nonnegative, we allow the edge weights to be negative.}

For expressiveness, as in GCN~\cite{kipf2016semi}, we first perform a learnable linear transformation of {each node's representation at the $l$-th layer}:
    $\widehat{\rmF^{(l)}} = \rmF^{(l)}\rmW^{(l)}+\rvb^{(l)}.$
Then, we define a sign function to be multiplied with the messages exchanged between neighbors.
To allow for backpropagation of the gradient information, we approximate the sign function with {cosine similarity. Denote $\rmS^{l}$ as the matrix which stores the sign information about the edges, defined as: 
$\rmS^{(l)}[i,j]=$\texttt{Cosine}($\rvf^{(l)}_i$, $\rvf^{(l)}_j$) if ($i \neq j$) \& ($v_j \in \gN_i$); 0 otherwise.}

In order to separate the contribution of similar neighbors (likely in the same class) from that of dissimilar neighbors (unlikely to be in the same class), we split $\rmS^{(l)}$ into a positive matrix $\rmS^{(l)}_{\text{pos}}$ and a negative matrix $\rmS^{(l)}_{\text{neg}}$. 
Thus, our proposed \method model learns a weighted combination of the self-representations,  the positive messages, and the negative messages: 

\vspace{-0.6cm}
{\small
\begin{equation}
    \rmF^{(l+1)} = \sigma\Big(\hat{\alpha^l} (\hat{\beta_0^l} \colorbox{gray!20}{$\widehat{\rmF^{(l)}}$} + \hat{\beta_1^l}\colorbox{gray!20}{\makebox(50,12){$(\rmS^{(l)}_{\text{pos}}\odot {\widehat{\tilde{\rmA}^{l}}})\widehat{\rmF^{(l)}}$}} + \hat{\beta_2^l}\colorbox{gray!20}{\makebox(50,12){$(\rmS^{(l)}_{\text{neg}}\odot {\widehat{\tilde{\rmA}^l}}) \widehat{\rmF^{(l)}}$}})\Big), \nonumber
\label{eq:sign_design}
\vspace{-0.2cm}
\end{equation}
}
where $\hat{\beta_0^l}$, $\hat{\beta_1^l}$ and $\hat{\beta_2^l}$ are scalars obtained by applying softmax to the learned scalars $\beta_0^l$, $\beta_1^l$ and $\beta_2^l$; the non-negative scaling factor $\hat{\alpha^l}=\text{softplus}(\alpha^l)$ 
is derived from the learned scalar $\alpha^l$; and $\sigma$ is the nonlinear function Elu. We note that we 
learn \textit{different} $\alpha$ and $\beta$ parameters \textit{per layer} for flexibility.
We also {require the} combined weights, $\hat{\alpha}^l \hat{\beta}^l_x$,  to be non-negative so that they do not negate the intended effect of the signed information.

\subsection{Decaying Aggregation}
Besides our two proposed mechanisms that are theoretically grounded in our analysis (\S~\ref{section: analysis}), we also incorporate into \method an existing design---decaying aggregation of messages---that empirically increases performance. 
However, we note that, even without this design, our GCN architecture still performs well under heterophily and is robust to oversmoothing (App. \S \ref{subsec:ablation}). 

Decaying aggregation was introduced in~\cite{chen2020simple} as a way to slow down the convergence rate of node representations. Inspired by this work, we modify the decaying function, $\hat{\eta}$, and integrate it to our \method model: 

{\small 
\begin{align}
\vspace{-1.2cm}
\rmF^{(l+1)} &= \colorbox{gray!20}{$\rmF^{(l)} + \hat{\eta}$}\bigg(\sigma\Big(\hat{\alpha^l} (\hat{\beta_0^l} \widehat{\rmF^{(l)}} 
+ \hat{\beta_1^l}(\rmS^{(l)}_{\text{pos}}\odot \tilde{\rmA} \odot \Tau^{(l)})\widehat{\rmF^{(l)}} \nonumber  \\
&+ \hat{\beta_2^l}(\rmS^{(l)}_{\text{neg}}\odot \tilde{\rmA} \odot \Tau^{(l)}) \widehat{\rmF^{(l)}})\Big)\bigg). 
\end{align}
}
\vspace{-0.5cm}

In practice, we found that the following decaying function works well: $\hat{\eta} \equiv \text{ln}(\frac{\eta}{l^k}+1), \text{iff}\,\,l\geq l_0; \hat{\eta} = 1, \text{otherwise}$. 
The hyperparameters $k$, $l_0$, $\eta$ are tuned on the validation set. 



\section{Experiments}
\label{section: experiments}
{We focus on the following four questions:}
{
({\bf Q1}) 
Compared to the baselines, how does \method perform on homophilous and heterophilous graphs?
({\bf Q2}) How robust is it against oversmoothing under homophily and heterophily?
({\bf Q3}) Is our node profiling effective in predicting the performance degradation of GCNs in heterophilous graphs?
  ({\bf Q4}) {How can we verify the correctness of our theorems about oversmoothing on real datasets?}}
{We provide an ablation study for 
our proposed edge correction mechanisms in
in App.~\S~\ref{subsec:ablation}.} 
\subsection{Experimental Setup}
\label{sec:exp_setup}
\noindent \textit{Datasets.} We evaluate the performance of our \method model and existing GNNs in node classification on various real-world datasets~\cite{tang2009social-fc,rozemberczki2019multiscale,sen2008collective, namata2012query,bojchevski2018deep, shchur2018pitfalls}. We provide their summary statistics 
in 
Table~\ref{tab:real-results}, where 
we compute the homophily level $h$ of a graph as the average of $h_i$ of all nodes $v_i\in\gV$. For all benchmarks, we use the 
feature vectors, class labels, and 10 random splits (48\%/32\%/20\% of nodes per class for train/validation/test\footnote{\cite{pei2019geom} claims that the ratios are 60\%/20\%/20\%, which is different from the actual data splits shared on GitHub.}) from~\cite{pei2019geom}. 

\vspace{0.1cm}
\noindent \textit{Baselines.} For baselines we use  
{\textbf{(1)} classic GNN models for node classification:
vanilla GCN~\cite{kipf2016semi},  GAT~\cite{velivckovic2017graph} and GraphSage~\cite{hamilton2017inductive}; 
\textbf{(2)} recent models tackling heterophily: Geom-GCN~\cite{pei2019geom}, H2GCN~\cite{zhu2020beyond}, FAGCN~\cite{bo2021beyond} and {GPRGNN~\cite{chien2021adaptive}}; 
\textbf{(3)} {models tackling oversmoothing:} {PairNorm~\cite{zhao2019pairnorm}} and GCNII~\cite{chen2020simple} (state-of-the-art);
and 
\textbf{(4)} 2-layer MLP (with dropout and Elu non-linearity).} For GCN, PairNorm, Geom-GCN, GCNII, H2GCN, and GPRGNN, we use the original codes provided by the authors. For GAT, we use the code from a well-accepted Github repository\footnote{https://github.com/Diego999/pyGAT}. For GraphSage and FAGCN, we report the results from \cite{zhu2020beyond, chen2022memory}, which uses the same data and splits. For the baselines that have multiple variants (Geom-GCN, GCNII, H2GCN), we choose the best variant per dataset and denote them as [model]*. {We give the hyperparameters in App.~\ref{app:hyperparameter}.}

\noindent \textit{Machine.} We ran our experiments on Nvidia V100 GPU.
\setlength{\tabcolsep}{4pt}
\begin{table*}[h]
    \centering
    \vspace{-0.2cm}
    \caption{
    Real data: mean accuracy $\pm$ stdev over different data splits. Per GNN model, we report the best performance  across different layers. Best model per benchmark highlighted in gray. The ``$^\dagger$'' results (GraphSAGE) are obtained from~~\protect\citep{zhu2020beyond}. 
    }
    \label{tab:real-results}
    \vspace{-0.3cm}
    \begin{adjustbox}{width=.9\textwidth}
    \begin{tabular}{lcccccc  ccc c c} 
    \toprule
       &  \texttt{\bf Texas}           &   \texttt{\bf Wisconsin}           &   \texttt{\bf Actor}            &   \texttt{\bf Squirrel}   &   \texttt{\bf Chameleon} & \texttt{\bf Cornell}        &   \texttt{\bf Citeseer}           &   \texttt{\bf Pubmed}            &   \texttt{\bf Cora} &&  \\
          \textbf{Hom.\ level} $h$ & \textbf{0.11} & \textbf{0.21} & \textbf{0.22} & \textbf{0.22} & \textbf{0.23} & \textbf{0.3} & \textbf{0.74} & \textbf{0.8} & \textbf{0.81} && {\multirow{4}{*}{\rotatebox[origin=c]{90}{\textbf{Avg Rank}}}}\\
		\textbf{\#Nodes} & 183 & 251 & 7,600 & 5,201 & 2,277 & 183 & 3,327 & 19,717 & 2,708 && \\
		\textbf{\#Edges} & 295 & 466 & 26,752 & 198,493 & 31,421 & 280 & 4,676 & 44,327 & 5,278 && \\
		\textbf{\#Classes} & 5 & 5 & 5 & 5 & 5 & 5 & 7 & 3 & 6 && \\ 
    \midrule
       {\method (\textbf{ours})} & \cellcolor{gray!15}$84.86{\scriptstyle\pm4.55}$ & $86.86{\scriptstyle\pm3.29}$ &
       \cellcolor{gray!15}$37.54{\scriptstyle\pm1.56}$ & \cellcolor{gray!15}$55.17{\scriptstyle\pm1.58}$ & \cellcolor{gray!15}$71.14{\scriptstyle\pm1.84}$ &  \cellcolor{gray!15}$85.68{\scriptstyle\pm6.63}$ &  $77.14{\scriptstyle\pm1.45}$ & $89.15{\scriptstyle\pm0.37}$ & $87.95{\scriptstyle\pm1.05}$ && \cellcolor{gray!15}1.78 \\
       {GPRGNN} & $78.38{\scriptstyle\pm4.36}$ & $82.94{\scriptstyle\pm4.21}$ & $34.63{\scriptstyle\pm1.22}$ & $31.61{\scriptstyle\pm1.24}$ & $46.58{\scriptstyle\pm1.71}$ & $80.27{\scriptstyle\pm8.11}$ &  $77.13{\scriptstyle\pm1.67}$ & $87.54{\scriptstyle\pm0.38}$ & $87.95{\scriptstyle\pm1.18}$ && 5.56 \\
       {H2GCN*} & $84.86{\scriptstyle\pm7.23}$ & \cellcolor{gray!15}$87.65{\scriptstyle\pm4.98}$ & $35.70{\scriptstyle\pm1.00}$ & $36.48{\scriptstyle\pm1.86}$ & $60.11{\scriptstyle\pm2.15}$ & $82.70{\scriptstyle\pm5.28}$ &  $77.11{\scriptstyle\pm1.57}$ & $89.49{\scriptstyle\pm0.38}$ & $87.87{\scriptstyle\pm1.20}$ && 3.89\\
       {GCNII*} & $77.57{\scriptstyle\pm3.83}$ & $80.39{\scriptstyle\pm3.4}$ & $37.44{\scriptstyle\pm1.30}$ & $38.47{\scriptstyle\pm1.58}$ & $63.86{\scriptstyle\pm3.04}$ &
       $77.86{\scriptstyle\pm3.79}$ &
       $77.33{\scriptstyle\pm1.48}$ &
       \cellcolor{gray!15}$90.15{\scriptstyle\pm0.43}$ &
       \cellcolor{gray!15}$88.37{\scriptstyle\pm1.25}$ && 3.56 \\
       {Geom-GCN*} & $66.76{\scriptstyle\pm2.72}$ & $64.51{\scriptstyle\pm3.66}$ & $31.59{\scriptstyle\pm1.15}$ & $38.15{\scriptstyle\pm0.92}$ & $60.00{\scriptstyle\pm2.81}$ & $60.54{\scriptstyle\pm3.67}$ &  \cellcolor{gray!15}$78.02{\scriptstyle\pm1.15}$ & $89.95{\scriptstyle\pm0.47}$ & $85.35{\scriptstyle\pm1.57}$ && 6.11 \\
       {PairNorm} & $60.27{\scriptstyle\pm4.34}$ & $48.43{\scriptstyle\pm6.14}$ &
       $27.40{\scriptstyle\pm1.24}$ & $50.44{\scriptstyle\pm2.04}$ & $62.74{\scriptstyle\pm2.82}$ &  $58.92{\scriptstyle\pm3.15}$ &  $73.59{\scriptstyle\pm1.47}$ & $87.53{\scriptstyle\pm0.44}$ & $85.79{\scriptstyle\pm1.01}$ && 7.78 \\
       {GraphSAGE$^\dagger$} & $82.43{\scriptstyle\pm6.14}$ & $81.18{\scriptstyle\pm5.56}$ & $34.23{\scriptstyle\pm0.99}$ & $41.61{\scriptstyle\pm0.74}$ & $58.73{\scriptstyle\pm1.68}$ & $75.95{\scriptstyle\pm5.01}$ &  $76.04{\scriptstyle\pm1.30}$ & $88.45{\scriptstyle\pm0.50}$ & $86.90{\scriptstyle\pm1.04}$ &&  5.78 \\
	   {GCN} & $55.14{\scriptstyle\pm5.16}$ & $51.76{\scriptstyle\pm3.06}$ &  $27.32{\scriptstyle\pm1.10}$ & $53.43{\scriptstyle\pm2.01}$ &
	   $64.82{\scriptstyle\pm2.24}$ & 
	   $60.54{\scriptstyle\pm5.3}$ &  $76.50{\scriptstyle\pm1.36}$ & $88.42{\scriptstyle\pm0.5}$ & $86.98{\scriptstyle\pm1.27}$ &&  6.56 \\
	   {GAT} & $52.16{\scriptstyle\pm6.63}$ & $49.41{\scriptstyle\pm4.09}$ & $27.44{\scriptstyle\pm0.89}$ & $40.72{\scriptstyle\pm1.55}$ & $60.26{\scriptstyle\pm2.5}$ & $61.89{\scriptstyle\pm5.05}$ &  $76.55{\scriptstyle\pm1.23}$ & $86.33{\scriptstyle\pm0.48}$ & $87.30{\scriptstyle\pm1.10}$ &&  7.22 \\
	   {MLP} & $80.81{\scriptstyle\pm4.75}$ & $85.29{\scriptstyle\pm3.31}$ & $36.53{\scriptstyle\pm0.70}$ & $28.77{\scriptstyle\pm1.56}$ & $46.21{\scriptstyle\pm2.99}$ & $81.89{\scriptstyle\pm6.40}$ &  $74.02{\scriptstyle\pm1.90}$ & $87.16{\scriptstyle\pm0.37}$ & $75.69{\scriptstyle\pm2.00}$ &&  6.78 \\
	   \bottomrule
    \end{tabular}
    \end{adjustbox}
\end{table*}

\setlength{\tabcolsep}{3pt}
\begin{table*}[h]
\centering
\caption{Model performance for different layers: mean accuracy $\pm$ stdev over different data splits. Per dataset and GNN model, we also report the layer at which the best performance (given in Table~\ref{tab:real-results}) is achieved. `OOM': out of memory; `INS': numerical instability. For larger font, refer to Table~\ref{app-tab:table 2} in the Appendix.}
\label{tab:oversmoothing}
\vspace{-0.15cm}
\begin{adjustbox}{width=1\linewidth}
{\footnotesize
\begin{tabular}{r c ccccccccccccccc}
\toprule
                                      \textbf{Layers} 
                                      && \textbf{2}           & \multicolumn{1}{c}{\textbf{4}} & \multicolumn{1}{c}{\textbf{8}} & \multicolumn{1}{c}{\textbf{16}} & \multicolumn{1}{c}{\textbf{32}} & \multicolumn{1}{c}{\textbf{64}} &
                                      \multicolumn{1}{c}{\textbf{Best}}
                                      &&  \multicolumn{1}{c}{\textbf{2}}           & \multicolumn{1}{c}{\textbf{4}} & \multicolumn{1}{c}{\textbf{8}} & \multicolumn{1}{c}{\textbf{16}} & \multicolumn{1}{c}{\textbf{32}} & \multicolumn{1}{c}{\textbf{64}}  &
                                      \multicolumn{1}{c}{\textbf{Best}} \\ \cmidrule{1-1} \cmidrule{3-9} \cmidrule{11-17} 
&& \multicolumn{7}{c}{\textbf{Cora} ($h$=0.81)}     & & \multicolumn{7}{c}{\textbf{Citeseer} ($h$=0.74)}                                                                         \\ 
\cline{3-9} \cline{11-17}
                                    
                                        \multicolumn{1}{l}{GGCN (\textbf{ours})}    &&   
                                        $87.00{\scriptstyle\pm1.15}$         & {$ 87.48{\scriptstyle\pm1.32}$}           & {$ 87.63{\scriptstyle\pm1.33}$}           & {$ 87.51{\scriptstyle\pm1.19}$}            & 
                                        {$ 87.95{\scriptstyle\pm1.05}$}            & {$ 87.28{\scriptstyle\pm1.41}$}            & 
                                        32
                                           & &  
                                        $76.83{\scriptstyle\pm1.82}$ & $76.77{\scriptstyle\pm1.48}$ & $76.91{\scriptstyle\pm1.56}$ & $76.88{\scriptstyle\pm1.56}$& $76.97{\scriptstyle\pm1.52}$ & $76.65{\scriptstyle\pm1.38}$ & 
                                        10 \\ 
                                        
                            \multicolumn{1}{l}{GPRGNN}    &&  
                                      \multicolumn{1}{c}{$87.93{\scriptstyle\pm1.11}$}      &        $87.95{\scriptstyle\pm1.18}$           & \multicolumn{1}{c}{$87.87{\scriptstyle\pm1.41}$}           & \multicolumn{1}{c}{$87.26{\scriptstyle\pm1.51}$}            & \multicolumn{1}{c}{$87.18{\scriptstyle\pm1.29}$}            & \multicolumn{1}{c}{$87.32{\scriptstyle\pm1.21}$} & 4 &   &                   
                                      $77.13{\scriptstyle\pm1.67}$  & $77.05{\scriptstyle\pm1.43}$&  $77.09{\scriptstyle\pm1.62}$&  $76.00{\scriptstyle\pm1.64}$&  $74.97{\scriptstyle\pm1.47}$&  $74.41{\scriptstyle\pm1.65}$
                            & 2
                            \\
                            \multicolumn{1}{l}{H2GCN*} &&  
                                      $87.87{\scriptstyle\pm1.20}$  &   $86.10{\scriptstyle\pm1.51}$&   $86.18{\scriptstyle\pm2.10}$&    OOM                    &       OOM                     &      OOM                      &     2
                                      & &    
                                    $76.90{\scriptstyle\pm1.80}$  & 
                                      $76.09{\scriptstyle\pm1.54}$  & 
                                      $74.10{\scriptstyle\pm1.83}$  &  
                                      OOM     &  
                                      OOM     &  
                                      OOM     &     1
                                      
                        \\      
                                     \multicolumn{1}{l}{GCNII*}    &&  
                                      $85.35{\scriptstyle\pm1.56}$                & {$85.35{\scriptstyle\pm1.48}$}           & {$86.38{\scriptstyle\pm0.98}$}           & {$87.12{\scriptstyle\pm1.11}$}            & {$87.95{\scriptstyle\pm1.23}$}            & {$88.37{\scriptstyle\pm1.25}$}      &       
                                      64  & &
                                      $75.42{\scriptstyle\pm1.78}$  & $75.29{\scriptstyle\pm1.90}$&  $76.00{\scriptstyle\pm1.66}$&  $76.96{\scriptstyle\pm1.38}$&  $77.33{\scriptstyle\pm1.48}$&  $77.18{\scriptstyle\pm1.47}$ & 32
                                      \\  
                                      
                        \multicolumn{1}{l}{PairNorm}    &&   
                                        $ 85.79{\scriptstyle\pm1.01}$         & \multicolumn{1}{c}{$ 85.07{\scriptstyle\pm0.91}$}    &        \multicolumn{1}{c}{$ 84.65{\scriptstyle\pm1.09}$}     &       \multicolumn{1}{c}{$ 82.21{\scriptstyle\pm2.84}$}      &     \multicolumn{1}{c}{$ 60.32{\scriptstyle\pm8.28}$}       &    \multicolumn{1}{c}{$ 44.39{\scriptstyle\pm5.60}$}   & 2 & & 
                                        $73.59{\scriptstyle\pm1.47}$ & $72.62{\scriptstyle\pm1.97}$ & $72.32{\scriptstyle\pm1.58}$ & $59.71{\scriptstyle\pm15.97}$& $27.21{\scriptstyle\pm10.95}$ & $23.82{\scriptstyle\pm6.64}$ & 2                                        \\               
                        \multicolumn{1}{l}{Geom-GCN*} &&
                          $85.35{\scriptstyle\pm1.57}$&   $21.01{\scriptstyle\pm2.61}$&   $13.98{\scriptstyle\pm1.48}$&   $13.98{\scriptstyle\pm1.48}$&   $13.98{\scriptstyle\pm1.48}$&   $13.98{\scriptstyle\pm1.48}$&   
                          2
                         &  & 
                         $78.02{\scriptstyle\pm1.15}$&  $23.01{\scriptstyle\pm1.95}$&   $7.23{\scriptstyle\pm0.87}$&   $7.23{\scriptstyle\pm0.87}$&   $7.23{\scriptstyle\pm0.87}$&   $7.23{\scriptstyle\pm0.87}$
                         &   2 
                         \\ 
                        \multicolumn{1}{l}{GCN} &&
                        $86.98{\scriptstyle\pm1.27}$ & $83.24{\scriptstyle\pm1.56}$ & $31.03{\scriptstyle\pm3.08}$ &
                        $31.05{\scriptstyle\pm2.36}$ & $30.76{\scriptstyle\pm3.43}$ & $31.89{\scriptstyle\pm2.08}$ & 2
                         &  & 
                        $76.50{\scriptstyle\pm1.36}$ & $64.33{\scriptstyle\pm8.27}$ & $24.18{\scriptstyle\pm1.71}$ &
                               $23.07{\scriptstyle\pm2.95}$ & $25.3{\scriptstyle\pm1.77}$ & $24.73{\scriptstyle\pm1.66}$ & 2  
                         \\              
                        
                         \multicolumn{1}{l}{GAT} &&
                         $87.30{\scriptstyle\pm1.10}$ & $86.50{\scriptstyle\pm1.20}$ & $84.97{\scriptstyle\pm1.24}$ &
                              INS  & INS & INS & 2
                         &  & 
                        $76.55{\scriptstyle\pm1.23}$ &$75.33{\scriptstyle\pm1.39}$&    $66.57{\scriptstyle\pm5.08}$         &    INS                             &    INS                             &       INS                          & 2  
                         \\
                         
                         \midrule 
&& \multicolumn{7}{c}{\textbf{Cornell ($h$=0.3)}} && \multicolumn{7}{c}{\textbf{Chameleon ($h$=0.23)}}                                                                         \\ 
\cline{3-9} \cline{11-17}
                                    
                        \multicolumn{1}{l}{GGCN (\textbf{ours})}    &&   
                        $83.78{\scriptstyle\pm6.73}$ & $83.78{\scriptstyle\pm6.16}$ & $84.86{\scriptstyle\pm5.69}$ &
                        $83.78{\scriptstyle\pm6.73}$ & 
                        $83.78{\scriptstyle\pm6.51}$ & 
                        $84.32{\scriptstyle\pm5.90}$ & 
                               6
                        & &  
                        $70.77{\scriptstyle\pm1.42}$ & $69.58{\scriptstyle\pm2.68}$ & $70.33{\scriptstyle\pm1.70}$ & $70.44{\scriptstyle\pm1.82}$& $70.29{\scriptstyle\pm1.62}$ & $70.20{\scriptstyle\pm1.95}$ & 
                        5                
                        \\ 
            \multicolumn{1}{l}{GPRGNN}    && $76.76{\scriptstyle\pm8.22}$ & $77.57{\scriptstyle\pm7.46}$ & $80.27{\scriptstyle\pm8.11}$ &
                              $78.38{\scriptstyle\pm6.04}$ & $74.59{\scriptstyle\pm7.66}$ & $70.00{\scriptstyle\pm5.73}$    & 8 & &
$46.58{\scriptstyle\pm1.771}$ & $45.72{\scriptstyle\pm3.45}$ & $41.16{\scriptstyle\pm5.79}$ & $39.58{\scriptstyle\pm7.85}$& $35.42{\scriptstyle\pm8.52}$ & $36.38{\scriptstyle\pm2.40}$ & 2
            \\
            \multicolumn{1}{l}{H2GCN*} &&  
                        $81.89{\scriptstyle\pm5.98}$ & $82.70{\scriptstyle\pm6.27}$ & $80.27{\scriptstyle\pm6.63}$ &
                               OOM & OOM & OOM & 1  
                        & &    
                        $59.06{\scriptstyle\pm1.85}$ & $60.11{\scriptstyle\pm2.15}$ & OOM &                          OOM &                          OOM &                          OOM &                          4            
                        \\  
            
                        \multicolumn{1}{l}{GCNII*} && 
                        $67.57{\scriptstyle\pm11.34}$ & $64.59{\scriptstyle\pm9.63}$ & $73.24{\scriptstyle\pm5.91}$ &
                               $77.84{\scriptstyle\pm3.97}$ & $75.41{\scriptstyle\pm5.47}$ & $73.78{\scriptstyle\pm4.37}$ & 16   
                        & &
                        $61.07{\scriptstyle\pm4.10}$ & $63.86{\scriptstyle\pm3.04}$ & $62.89{\scriptstyle\pm1.18}$ & $60.20{\scriptstyle\pm2.10}$& $56.97{\scriptstyle\pm1.81}$ & $55.99{\scriptstyle\pm2.27}$ & 4  
                        \\ 
                        \multicolumn{1}{l}{PairNorm}    && $50.27{\scriptstyle\pm7.17}$ & $53.51{\scriptstyle\pm8.00}$ & $58.38{\scriptstyle\pm5.01}$ &
                              $58.38{\scriptstyle\pm3.01}$ & $58.92{\scriptstyle\pm3.15}$ & $58.92{\scriptstyle\pm3.15}$          & 32 & &
$62.74{\scriptstyle\pm2.82}$ & $59.01{\scriptstyle\pm2.80}$ & $54.12{\scriptstyle\pm2.24}$ & $46.38{\scriptstyle\pm2.23}$& $46.78{\scriptstyle\pm2.26}$ & $46.27{\scriptstyle\pm3.24}$ & 2
\\  
                        
                        \multicolumn{1}{l}{Geom-GCN*} &&
                        $60.54{\scriptstyle\pm3.67}$&  $23.78{\scriptstyle\pm11.64}$& $12.97{\scriptstyle\pm2.91}$&  $12.97{\scriptstyle\pm2.91}$&  $12.97{\scriptstyle\pm2.91}$&  $12.97{\scriptstyle\pm2.91}$&  
                        2  
                         &  & 
                        $60.00{\scriptstyle\pm2.81}$&  $19.17{\scriptstyle\pm1.66}$& $19.58{\scriptstyle\pm1.73}$&  $19.58{\scriptstyle\pm1.73}$&  $19.58{\scriptstyle\pm1.73}$&  $19.58{\scriptstyle\pm1.73}$
                        &  2 
                        \\
                         \multicolumn{1}{l}{GCN} &&
                        $60.54{\scriptstyle\pm5.30}$ & $59.19{\scriptstyle\pm3.30}$ & $58.92{\scriptstyle\pm3.15}$ &
                        $58.92{\scriptstyle\pm3.15}$ & 
                        $58.92{\scriptstyle\pm3.15}$ & 
                        $58.92{\scriptstyle\pm3.15}$ & 2
                               &  & 
                        $64.82{\scriptstyle\pm2.24}$ & $53.11{\scriptstyle\pm4.44}$ & $35.15{\scriptstyle\pm3.14}$ &
                               $35.39{\scriptstyle\pm3.23}$ & $35.20{\scriptstyle\pm3.25}$ & $35.50{\scriptstyle\pm3.08}$ & 2 
                         \\ 
                         \multicolumn{1}{l}{GAT} &&
                        $61.89{\scriptstyle\pm5.05}$ &      $58.38{\scriptstyle\pm4.05}$ &     $58.38{\scriptstyle\pm3.86}$                           
                        &   INS                    
                        &   INS                         
                        &   INS &  
                        2 
                         &  & 
                        $60.26{\scriptstyle\pm2.50}$ &   $48.71{\scriptstyle\pm2.96}$                             &  $35.09{\scriptstyle\pm3.55}$                              &      INS                           &    INS                             &   INS                              &  2 
                         
                        \\ \bottomrule
\end{tabular}
}
\end{adjustbox}
\vspace{-0.2cm}
\end{table*}

\subsection{(Q1) Performance Under Homophily \& Heterophily} 
Table~\ref{tab:real-results} provides the test accuracy of different GNNs on the supervised node classification task over datasets with varying homophily levels (arranged from low homophily to high homophily). {A graph's homophily level is the average of its nodes' homophily levels.}
We report the best performance of each model across different layers. 

\method performs the best in terms of average rank (1.78) across all datasets, which suggests its strong adaptability to graphs of various homophily levels. In particular, it achieves the highest accuracy in 5 out of 6 heterophilous graphs ({$h$ is low}). For datasets like Chameleon and Cornell,  \method enhances accuracy by around 6\% and 3\% compared to the second-best model. On homophily datasets (Citeseer, Pubmed, Cora), the accuracy of \method is within a 1\% difference of the best model. 

Our experiments highlight that MLP is a good baseline in heterophilous datasets. In heterophilous graphs, the models that are not specifically designed for heterophily usually perform worse than an MLP. Though H2GCN* is the second best model in heterophilous datasets, we can still see that in the Actor dataset, MLP performs better. {GPRGNN, FAGCN and Geom-GCN*}, which are specifically designed for heterophily,
achieve better performance than classic GNNs (GCN and GAT) in heterophilous datasets, but {do not show clear advantage over MLP}.
Our \method model is the only model that performs better than MLP across all the datasets.

In general, GNN models perform well in homophilous datasets. GCNII* performs the best, and \method, H2GCN*, GPRGNN and Geom-GCN* also achieve high performance.

\setlength{\tabcolsep}{4pt}
\begin{table}[h]
    \centering
    \caption{
    The percentage (\%) of nodes per case (Fig.~\ref{fig:concept} and Thm.~\ref{theorem: initial_stage}): case 1: $h_i\leq \frac{\rho}{1+\rho}$, case 2: $h_i>\frac{\rho}{1+\rho} \ \mathrm{\&}\  \ratioi \leq \frac{1}{(1+\rho)h_i-\rho}$ and case 3: otherwise. The dominant case per dataset is  in grey.
    }
    \label{tab:ratios}
    \vspace{-0.15cm}
    \begin{adjustbox}{width=\columnwidth}
    \begin{tabular}{lcccccc  ccc c} 
    \toprule
       &  \texttt{\bf Texas}           &   \texttt{\bf Wisconsin}           &   \texttt{\bf Actor}            &   \texttt{\bf Squirrel}   &   \texttt{\bf Chameleon} & \texttt{\bf Cornell}        &   \texttt{\bf Citeseer}           &   \texttt{\bf Pubmed}            &   \texttt{\bf Cora} \\
          \textbf{Hom.} $h$ & \textbf{0.11} & \textbf{0.21} & \textbf{0.22} & \textbf{0.22} & \textbf{0.23} & \textbf{0.3} & \textbf{0.74} & \textbf{0.8} & \textbf{0.81} \\
    \midrule
       {Case 1} & \cellcolor{gray!15}$87.43$ & \cellcolor{gray!15}$78.49$ & \cellcolor{gray!15}
       $63.49$ & \cellcolor{gray!15}$47.74$ & \cellcolor{gray!15}$48.48$ &  \cellcolor{gray!15}$79.23$ &  $18.33$ & $14.80$ & $6.50$ \\
       {Case 2} & $8.74$ & $15.93$ & $29.53$ & \cellcolor{gray!15}$50.14$ & \cellcolor{gray!15} $48.84$ & $6.01$ &  $25.58$ & $33.75$ & $32.98$ \\
       {Case 3} & $3.83$ & $5.58$ & $6.99$ & $2.11$ & $2.68$ & $14.75$ & \cellcolor{gray!15} $56.09$ & \cellcolor{gray!15}$51.45$ & \cellcolor{gray!15}$60.52$ \\
	   \bottomrule
    \end{tabular}
    \end{adjustbox}
    \vspace{-0.5cm}
\end{table}

\subsection{(Q2) Oversmoothing}
We also test how robust the models are to oversmoothing. To this end, 
we measure the supervised node classification accuracy for 2 to 64 layers. Table~\ref{tab:oversmoothing} presents the results for two homophilous datasets (top) and two heterophilous datasets (bottom). Per model, we also report the layer at which the best performance is achieved (column `Best'). 

According to Table~\ref{tab:oversmoothing}, {\method and GCNII* 
achieve \textbf{increase} in accuracy when stacking more layers in four datasets, while GPRGNN and PairNorm  exhibit robustness against oversmoothing.} 
Models that are not designed for oversmoothing have various issues. The performance of GCN and Geom-GCN* drops rapidly as the number of layers grows; H2GCN* requires concatenating all the intermediate outputs and quickly reaches memory capacity; GAT's attention mechanism also has high memory requirements. We also find that GAT needs careful initialization when stacking many layers as it may suffer from numerical instability in sparse tensor operations. 

In general, models like \method, GCNII*, and GPRGNN that perform well under heterophily usually exhibit higher resilience against oversmoothing. One exception is Geom-GCN*, which suffers more than GCN. {This model incorporates structurally similar nodes 
into each node's neighborhood; this design may benefit Geom-GCN* in the shallow layers as the node degrees increase.} 
However, as we point out in Thm.~\ref{theorem: developing stage}, when the effective homophily is low, higher degrees are harmful. If the {structurally similar nodes} 
introduce lower homophily levels, 
their performance will rapidly degrade once the effective homophily is lower than $\frac{\rho_{i}^{l}}{1+\rho_{i}^{l}}$. On the other hand, \method \textit{virtually} changes the degrees thanks to the structure-based correction mechanism (\S~\ref{subsec: degree}),  
and, in practice, this design has \textbf{positive} impact on its robustness to oversmoothing.

\subsection{(Q3) Effectiveness of Node Profiling} 
{In Table~\ref{tab:ratios}, we provide results both for the three cases that our theory determines and for 
the frequently-used graph-level homophily $h$, which is defined as the average of the nodes’ homophily levels.
We observe that our node profiling better explains the performance of GCN shown in Table~\ref{tab:real-results}. In the Texas, Wisconsin and Actor datasets, only case 1 dominates; this case hurts the performance most, which explains why GCN performs worse than most methods. In the Squirrel and Chameleon datasets, both case 1 and case 2 dominate, so GCN yields reasonable results though the graph-level homophily is still low. Using the three cases to analyze the datasets is a better metric, because they align better with  GCN's performance.}


\begin{figure}[t]
\centering
\vspace{2cm}
\includegraphics[scale=0.13,bb=900 300 380 180]{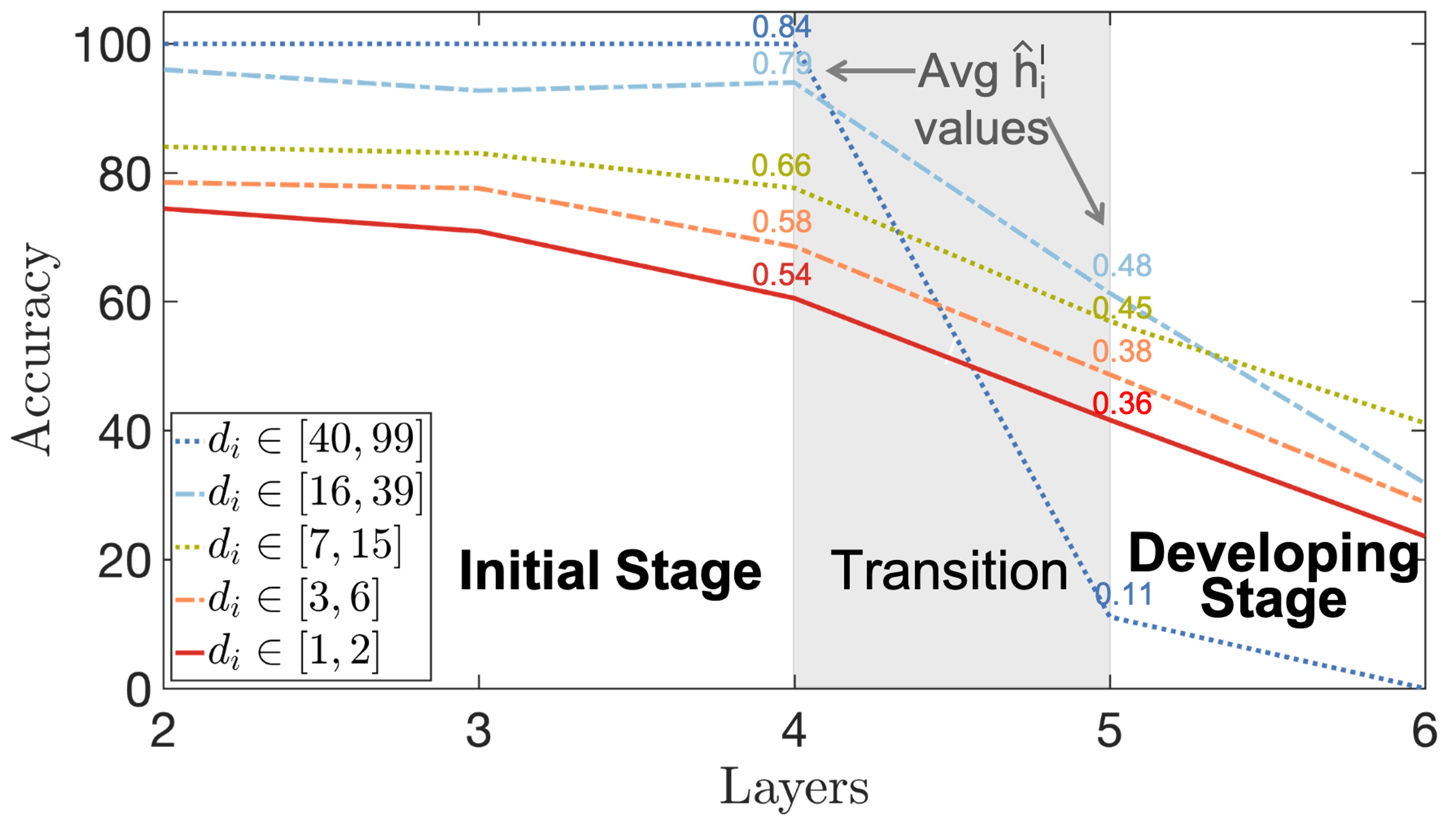}
\vspace{1.2cm}
\caption{\small Accuracy of nodes grouped by degree $d_i$ on Citeseer. Initial stage: when mean effective homophily $\heterophilyil$ (ratio of a node's neighbors in the same class--\S~\ref{sec:developing-stage}) is high, the accuracy increases as the degree increases. 
Developing stage: when $\heterophilyil$ is low, the accuracy of high-degree nodes drops more sharply. 
}
\label{fig:phase_transition}
\vspace{-0.4cm}
\end{figure}

\subsection{(Q4) Empirical Verification of the Two Stages} 
\label{exp:case study}


Using the vanilla GCN model~\cite{kipf2016semi}, we validate our theorems by measuring the test accuracy and effective homophily for different node degrees (binned logarithmically) on real datasets. 
We estimate the effective homophily as the portion of the same-class neighbors that are correctly classified \textit{before} the last propagation. 
Figure~\ref{fig:phase_transition} shows the results for Citeseer. 
In the initial stage (high $\heterophilyil$), the accuracy increases with the degree, but in the developing stage, the trend changes, with high-degree nodes being impacted the most, as predicted by our theorems.
\section{Related Work}
\label{section:related_work}
\textbf{Graph Convolutional Neural Networks.}  Early on, \citep{defferrard2016convolutional} proposed a {GCN model} that combines spectral filtering of graph signals and non-linearity for supervised node classification.
The scalability and numerical stability of GCNs was later improved with a localized first-order approximation of spectral graph convolutions proposed in 
\citep{kipf2016semi}.
\citep{velivckovic2017graph} proposes the first graph attention network to improve neighborhood aggregation.
Many more GCN variants have been proposed for different applications such as: computer vision~\citep{satorras2018few}, social science~\citep{li2019encoding}, biology~\citep{yan2019groupinn},  algorithmic tasks~\citep{velivckovic2020pointer, yan2020neural}, and inductive  classification~\citep{hamilton2017inductive}. Concurrent work~\citep{baranwal2021graph} provides theoretical analysis on the linear separability of graph convolution but does not provide effective strategies to increase the separability.

\textbf{Oversmoothing.}
The oversmoothing problem was first discussed in~\citep{li2018deeper}, which proved that by repeatedly applying Laplacian smoothing, the representations of nodes within each connected component of the graph converge to the same value.
Since then, various {empirical} solutions have been proposed:
residual connections and dilated convolutions~\citep{li2019deepgcns};
skip links~\citep{xu2018representation};
new normalization strategies~\citep{zhao2019pairnorm};
edge dropout~\citep{rong2019dropedge};
and a new model that even increases performance as more layers are stacked~\citep{chen2020simple}. {Some recent works provide theoretical analyses:} 
\citep{oono2019graph} showed that a $k$-layer renormalized graph convolution with a residual link simulates a lazy random walk 
and~\citep{chen2020simple} proved that the convergence rate is related to the spectral gap of the graph. 

\textbf{Heterophily \& GCNs.}
Heterophily has recently been recognized as an important issue for GCNs. It is first outlined in the context of GCNs in~\citep{pei2019geom}. 
\citep{zhu2020beyond} identified a set of effective designs that allow GCNs to generalize to challenging heterophilous settings, and 
\citep{zhu2020graph} introduced a new GCN model that leverages ideas from belief propagation~\citep{linbp}. 
Though recent work~\citep{chen2020simple} focused on solving the oversmoothing problem, it also empirically showed improvement on heterophilous datasets;  
these empirical observations formed the basis of our work.
Finally, \citep{chien2021adaptive} recently proposed a PageRank-based model that performs well under heterophily and  alleviates the oversmoothing problem. {However, they view the two problems independently and analyze their model via an asymptotic spectral perspective. Our work studies the representation dynamics and unveils the connections between the oversmoothing and heterophily problems theoretically and empirically. As we demonstrate with GGCN, addressing both issues in a principled manner provides superior performance across a variety of datasets.}


\section{Conclusion}
\label{section: conclusion}
Our work provides the first theoretical and empirical analysis that unveils the connections between the oversmoothing and heterophily problems. 
{By analyzing the statistical change of the node representations after the graph convolution, we identified two causes, i.e., the relative degree of a node compared to its neighbors and the level of heterophily in its neighborhood, which influence {the movements of node representations and lead to a higher misclassification rate}. Based on our new, unified theoretical perspective, we obtained three important insights: (1) Nodes with high heterophily {tend to be misclassified} after graph convolution; (2) Even with low heterophily, low-degree nodes can trigger a pseudo-heterophily situation that explains oversmoothing. {(3) Signed edges (instead of only positive edge weights) help alleviate the heterophily and oversmoothing problems.}}
Based on these insights, we designed a generalized model, \method, that {addresses the identified  causes using structure- and feature-based correction mechanisms.} 
Though other designs may also address these two problems, 
our work points out two effective directions that are theoretically grounded (\S~\ref{section: analysis}). 
{In summary, our research suggests it is beneficial to study these two 
problems jointly; this leads to architectural insights that can improve the learned representations of GNN models across a variety of domains.}

\bibliography{nee_bib.bib}
\bibliographystyle{icml2022}

\newpage
\appendix
\onecolumn
\newpage
\counterwithin{figure}{section}
\counterwithin{table}{section}

\section{Detailed Proofs of Theorems in \S~\ref{section: analysis}}
\label{app:proof}
\subsection{Proof of Theorem~\ref{theorem: initial_stage}}
\label{app:proof1}
\begin{proof}

{The node representations at the first layer are given by:}
\begin{equation}
\rvf_i^{(1)}=\frac{\rvf_i^{(0)}}{d_i+1} + \sum_{j \in \gN_i}{\frac{\rvf_j^{(0)}}{\sqrt{d_i+1} \cdot \sqrt{d_j+1}}}.
\label{eq:1}
\end{equation}

Without loss of generality, we assume node $v_i$ is in the first class $\gV_1$.
Then, we can express the conditional expectation of
  $\rvf_i^{(1)}$ as:
\begin{equation}
\E_{\rmA, \{y_i\}, \{\rvf_i^{(0)}\}|d_i, v_i \in \gV_1}(\rvf_i^{(1)}|v_i \in \gV_1, d_i) = \E_{\rmA|d_i, v_i \in \gV_1}(\E_{\{y_i\}, \{\rvf_i^{(0)}\}|\rmA,v_i \in \gV_1}(\rvf_i^{(1)}|v_i \in \gV_1, \rmA))
\label{eq:2}
\end{equation}
{Recall that $\rmA$ is the graph adjacency matrix; the first expectation is taken over the randomness of ground truth labels and initial input features, and the second expectation is taken over the randomness of graph structure ($\rmA$) given the degree of node $v_i$ and its label.} 
\begin{equation}
\begin{split}
& \E_{\{y_i\}, \{\rvf_i^{(0)}\}|\rmA,v_i \in \gV_1}(\rvf_i^{(1)}|v_i \in \gV_1, \rmA) \\=& \E_{\{y_i\}, \{\rvf_i^{(0)}\}|\rmA,v_i \in \gV_1}\left (\frac{\rvf_i^{(0)}}{d_i+1}|v_i \in \gV_1, \rmA \right ) + \sum_{j \in \gN_i}\E_{\{y_i\}, \{\rvf_i^{(0)}\}|\rmA,v_i \in \gV_1}\Bigg (\frac{\rvf_j^{(0)}}{\sqrt{d_i+1} \cdot \sqrt{d_j+1}}|v_i \in \gV_1, \rmA \Bigg) \\
=& \frac{\rvmu}{d_i+1} + \sum_{j \in \gN_i} \Bigg(\E_{\{y_i\}, \{\rvf_i^{(0)}\}|\rmA,v_i, v_j \in \gV_1}\left(\frac{\rvf_j^{(0)}}{\sqrt{d_i+1} \cdot \sqrt{d_j+1}}|\rmA, v_i, v_j \in \gV_1\right) \cdot \sP(v_j \in \gV_1|\rmA, v_i \in \gV_1)\\
+& \E_{\{y_i\}, \{\rvf_i^{(0)}\}|\rmA,v_i \in \gV_1,v_j \in \gV_2}\left(\frac{\rvf_j^{(0)}}{\sqrt{d_i+1} \cdot \sqrt{d_j+1}}|\rmA, v_i \in \gV_1, v_j \in \gV_2\right) \cdot \sP(v_j \in \gV_2|\rmA, v_i \in \gV_1)\Bigg)
\end{split}
\end{equation}
Due to our assumption (3) in Section~\ref{section: analysis}, $\rmA$ is independent of $\{y_i\}$. Thus, for \underline{$\forall v_j \in \gN_i$}, we have:
\begin{equation}
    \sP(v_j \in \gV_1|\rmA, v_i \in \gV_1)=\sP(v_j \in \gV_1|v_i \in \gV_1)=\sP(y_j=y_i|y_i=1)
\end{equation}
Given that $\{y_i=y_j|v_j\in\gN_i\}$ is independent of $y_i$ (assumption (2) in Section~\ref{section: analysis}) and $h_i = \sP(y_i=y_j|v_j \in \gN_i)$, therefore, when $v_j\in\gN_i$, $\sP(v_j \in \gV_1|\rmA, v_i \in \gV_1)=\sP(y_j=y_i|y_i=1)=\sP(y_j=y_i)=h_i$. Similarly, we have: $\sP(v_j \in \gV_2|\rmA, v_i \in \gV_1)=1-h_i$, given $v_j\in\gN_i$.

Given that $\rmA$ and $\{\rvf_i^{(0)}\}$ are independent, and $\rvf_j^{(0)}$ and $y_i$ are independent conditioned on $y_j$ (assumption (3)), $$\E_{\{y_i\}, \{\rvf_i^{(0)}\}|\rmA,v_i, v_j \in \gV_1}(\rvf_j^{(0)})=\E_{\{y_i\}, \{\rvf_i^{(0)}\}|v_j \in \gV_1}(\rvf_j^{(0)})=\rvmu$$
Similarly, we have: $$\E_{\{y_i\}, \{\rvf_i^{(0)}\}|\rmA,v_i\in \gV_1, v_j \in \gV_2}(\rvf_j^{(0)})=\rho\rvmu$$
Thus, we have:
\begin{equation}
\begin{split}
%
\E_{\{y_i\}, \{\rvf_i^{(0)}\}|v_i \in \gV_1, \rmA}(\rvf_i^{(1)}|v_i \in \gV_1, \rmA) =& \frac{\rvmu}{d_i+1} + \sum_{j \in \gN_i}\left(\frac{h_i}{\sqrt{d_i+1} \cdot \sqrt{d_j+1}}\rvmu-\frac{(1-h_i)}{\sqrt{d_i+1} \cdot \sqrt{d_j+1}}\rho\rvmu\right)\\
%
=& \frac{\rvmu}{d_i+1} + \frac{((1+\rho)h_i-\rho)\rvmu}{\sqrt{d_i+1}}\sum_{j \in \gN_i}{\frac{1}{\sqrt{d_j+1}}}\\
=& \left(\frac{1}{d_i+1}+\frac{\sum_{j \in \gN_i}{\frac{1}{\sqrt{d_j+1}}}}{\sqrt{d_i+1}}((1+\rho)h_i-\rho)\right)\rvmu \\
=& \left(\frac{1}{d_i+1}+\frac{\sum_{j \in \gN_i}{\frac{\sqrt{d_i + 1}}{\sqrt{d_j+1}}}}{d_i+1}((1+\rho)h_i-\rho)\right)\rvmu \\
=& \left(\frac{1}{d_i+1}+\frac{d_i}{d_i+1}\frac{\sum_{j \in \gN_i}r_{ij}}{d_i}((1+\rho)h_i-\rho)\right)\rvmu.
\end{split}
\label{eq:3}
\end{equation}
Given that $\{y_i\}$ is independent of $\rmA$ (assumption (3)), we can obtain the following equation by combining Equation~(\ref{eq:2}) and Equation~(\ref{eq:3}):
\begin{equation}
\begin{split}
\E_{\rmA, \{y_i\}, \{\rvf_i^{(0)}\}|d_i, v_i \in \gV_1}(\rvf_i^{(1)}|v_i \in \gV_1, d_i) &= \frac{1}{d_i+1}+\frac{d_i}{d_i+1}\E_{\rmA|d_i, v_i \in \gV_1}\left(\frac{\sum_{j \in \gN_i}r_{ij}}{d_i}\right)\left((1+\rho)h_i-\rho\right)\rvmu  \\
&=\left(\frac{1 + ((1+\rho)h_i-\rho)d_i\ratioi}{d_i+1} \right)\rvmu \equiv \gamma_i^1 \rvmu.
\end{split}
\end{equation}
%
%
%
%
Define $\epsilon \equiv (1+\rho)h_i-\rho$, and we consider three cases: \textbf{(1)} $h_i\leq\frac{\rho}{1+\rho}$, \textbf{(2)} $h_i > \frac{\rho}{1+\rho} \ \mathbf{\&} \  \ratioi \leq \frac{1}{\epsilon}$, and \textbf{(3)} $h_i > \frac{\rho}{1+\rho} \ \mathbf{\&} \  \ratioi > \frac{1}{\epsilon}$.


\begin{itemize}
    \item 
\colorbox{gray!30}{\textbf{CASE 1: $\pmb{h_i\leq\frac{\rho}{1+\rho}}$}}

$\bullet$ \textbf{Upper Bound}

We have:
\begin{equation}
\gamma_i^1 \leq \frac{1}{d_i+1} \leq \frac{1}{2}.
\label{eq:5}
\end{equation}
$\bullet$  \textbf{Lower Bound}


When $h_i \leq \frac{\rho}{1+\rho}$, we have:
\begin{enumerate}
    \item $((1+\rho)h_i-\rho) \leq 0$
    \item $\frac{d_i}{d_i+1}$ is an increasing non-negative function of $d_i$
    \item $\frac{1}{d_i+1}$ is a decreasing function of $d_i$ 
\end{enumerate}
Since $\ratioi$ is an increasing non-negative function of $d_i$, $\frac{1 + ((1+\rho)h_i-\rho)d_i\ratioi}{d_i+1}$ is a decreasing function of $d_i$.

When $h_i=\frac{\rho}{1+\rho}$, $\frac{1 + ((1+\rho)h_i-\rho)d_i\ratioi}{d_i+1} = \frac{1}{d_i+1}$ and $0 < \frac{1}{d_i+1} \leq \frac{1}{2}$.

When $h_i<\frac{\rho}{1+\rho}$,
\begin{equation}
\begin{split}
    \frac{1 + ((1+\rho)h_i-\rho)d_i\ratioi}{d_i+1} &\leq \frac{((1+\rho)h_i-\rho)d_i\ratioi}{d_i+1}+\frac{1}{2} \\&\leq \frac{((1+\rho)h_i-\rho)\ratioi}{2}+\frac{1}{2}.
\end{split}
\end{equation}
And we know that:
\begin{equation}
\lim_{\ratioi \to \infty}{\frac{((1+\rho)h_i-\rho)\ratioi}{2}+\frac{1}{2}}=-\infty.
\end{equation}
Thus,
\begin{equation}
\begin{split}
\lim_{\ratioi \to \infty}{\frac{1 + ((1+\rho)h_i-\rho)d_i\ratioi}{d_i+1}}=-\infty.
\end{split}
\label{eq:lower_bound}
\end{equation}



\vspace{0.2cm}
\item \colorbox{gray!30}{\textbf{CASE 2: $\pmb{h_i > \frac{\rho}{1+\rho} \;\;\; \mathbf{\&} \;\;\;  \ratioi \leq \frac{1}{\epsilon}}$}}

If $h_i >\frac{\rho}{1+\rho}$, $0 < \epsilon \leq 1$; if $\ratioi \leq \frac{1}{\epsilon}$, $0 < \epsilon \ratioi \leq 1$. Given 
\begin{equation}
\begin{split}
    \E(\rvf_i^{1}) = \left (\frac{1 + \epsilon d_i \ratioi}{d_i+1}\right )\rvmu,
\end{split}
\label{eq:6}
\end{equation}
we have: 
\begin{equation}
    {0<\frac{1}{d_i+1} < \gamma_i^1 \leq 1}.
\end{equation}


\vspace{0.2cm}
\item \colorbox{gray!30}{\textbf{CASE 3: $\pmb{h_i > \frac{\rho}{1+\rho} \;\;\; \mathbf{\&} \;\;\;  \ratioi > \frac{1}{\epsilon}}$}}

In this case, Equation~(\ref{eq:6}) still holds because $h_i>\frac{\rho}{1+\rho}$.

$\bullet$  \textbf{Lower Bound}

If $\ratioi > \frac{1}{\epsilon}$, then $\epsilon \ratioi > 1$, and therefore $\gamma_i^1 > 1$.

$\bullet$  \textbf{Upper Bound}

When $\epsilon>0$,
\begin{equation}
\begin{split}
    \frac{1 + \epsilon d_i\ratioi}{d_i+1} &> \frac{\epsilon d_i\ratioi}{d_i+1}\\&\geq \frac{\epsilon\ratioi}{2}.
\end{split}
\end{equation}
Because
\begin{equation}
    \lim_{\ratioi \to \infty}{\frac{\epsilon\ratioi}{2}}=\infty,
\end{equation}
we have:
\begin{equation}
    \lim_{\ratioi \to \infty}{\frac{1 + \epsilon d_i \ratioi}{d_i+1}}=\infty.
\end{equation}

Given that $\frac{d_i}{d_i+1}$ is an increasing non-negative function of $d_i$ and $\epsilon>0$, and that $\ratioi$ is an increasing non-negative function of $d_i$, $\frac{1 + \epsilon d_i\ratioi}{d_i+1}$ is an increasing function of $d_i$.

To sum it up, 
\begin{equation}
\gamma_i^1 \in 
\begin{cases}
  (-\infty, \frac{1}{2}], & \text{if}\ h_i\leq\frac{\rho}{1+\rho} \\
  (0, 1], & \text{if}\ h_i>\frac{\rho}{1+\rho} \ \mathrm{\&}\  \ratioi \leq \frac{1}{(1+\rho)h_i-\rho}\\
  (1, \infty), & \mathrm{otherwise}.
\end{cases}
\end{equation}

\end{itemize}

\vspace{0.5cm}
\textbf{Remarks: Special cases}

For different types of graphs and different nodes in the graph, $\ratioi$ might be different. For a \textbf{d-regular graph} whose nodes have a constant degree or a graph whose adjacency matrix is \textbf{row-normalized}, we will have:
\begin{equation}
\begin{split}
    \gamma_i^1 =& \left (\frac{1 + ((1+\rho)h_i-\rho)d_i\ratioi}{d_i+1}\right ) \\\leq & (\frac{1}{(d_i+1)}+\frac{d_i}{(d_i+1)}) = 1
\end{split}
\label{eq:7}
\end{equation}
Equality is achieved if and only if $h_i=1$. To note that, $h_i=1$ is not achievable for every node as long as there are more than one class. \textbf{Boundary nodes} will suffer most.

\end{proof}



\subsection{Proof of Theorem~\ref{theorem: developing stage}}
\label{app:proof2}
\begin{proof}
In the following proof, we use $\hat{\mathcal{N}_i^s}$ to refer to $\hat{\mathcal{N}_i^s}(\rmA, y_i, d_i)$ for conciseness.

The representations at the $(l+1)$-th layer are given by:
\begin{equation}
    \mathbf{f}_i^{(l+1)}=\frac{\mathbf{f}_i^{(l)}}{d_i+1} + \sum_{j \in \mathcal{N}_i}{\frac{\mathbf{f}_i^{(l)}}{\sqrt{d_i+1} \cdot \sqrt{d_j+1}}}
\end{equation}
\begin{equation}
\begin{split}
& \mathbb{E}_{\rmA,\{y_i\},\{\rvf_i^{(0)}\}|d_i,y_i}\Bigg(\mathbf{f}_i^{(l+1)}|d_i,y_i\Bigg) \\=& \mathbb{E}_{\rmA,\{y_i\},\{\rvf_i^{(0)}\}|d_i,y_i}\left (\frac{\mathbf{f}_i^{(l)}}{d_i+1}|d_i,y_i \right ) + \mathbb{E}_{\rmA,\{y_i\},\{\rvf_i^{(0)}\}|d_i,y_i}\left(\sum_{j \in \mathcal{N}_i}\frac{\mathbf{f}_j^{(l)}}{\sqrt{d_i+1} \cdot \sqrt{d_j+1}}|d_i,y_i \right) \\ =& \frac{\xi_i^l\E(\rvf^{(0)}_i|y_i)}{d_i+1} + \frac{1}{d_i+1}\mathbb{E}_{\rmA,\{y_i\},\{\rvf_i^{(0)}\}|d_i,y_i}\Bigg(\sum_{j \in \mathcal{N}_i}\frac{\mathbf{f}_j^{(l)}\sqrt{d_i+1}}{\sqrt{d_j+1}}|d_i, y_i\Bigg)
\end{split}
\label{eq:44}
\end{equation}
{In Section~\ref{section: analysis}, we assume that $\{d_j\}$ follow the same distribution and their joint distribution function is permutation-invariant, and $\{y_j\}$ and $\{\rvf_j^{l}\}$ also have this property. Thus, conditioned on $d_i$, $y_i$ and $\gN_i$,
the distribution of $\frac{\mathbf{f}_j^{(l)}\sqrt{d_i+1}}{\sqrt{d_j+1}}$  is the same for $\forall j \in \mathcal{N}_i$ (neighbors are indistinguishable) and we obtain:
\begin{eqnarray}
& &\mathbb{E}_{\rmA,\{y_i\},\{\rvf_i^{(0)}\}|d_i,y_i, \gN_i}\Bigg(\sum_{j \in \mathcal{N}_i}\frac{\mathbf{f}_j^{(l)}\sqrt{d_i+1}}{\sqrt{d_j+1}}|d_i, y_i, \gN_i\Bigg) \\&=& d_i\mathbb{E}_{\rmA,\{y_i\},\{\rvf_i^{(0)}\}|d_i,y_i, \gN_i, v_j \in \gN_i}\Bigg(\frac{\mathbf{f}_j^{(l)}\sqrt{d_i+1}}{\sqrt{d_j+1}}|d_i, y_i, \gN_i, v_j \in \gN_i\Bigg)\end{eqnarray}
\sloppy Given that the joint distribution of $\{d_j\}$, the joint distribution of $\{y_j\}$ and the joint distribution of $\{\rvf_j^{l}\}$ are permutation-invariant functions, for any $\gN_i$,
$\mathbb{E}_{\rmA,\{y_i\},\{\rvf_i^{(0)}\}|d_i,y_i, \gN_i}\Bigg(\sum_{j \in \mathcal{N}_i}\frac{\mathbf{f}_j^{(l)}\sqrt{d_i+1}}{\sqrt{d_j+1}}|d_i, y_i, \gN_i\Bigg)$ is the same. That is:
\begin{eqnarray}
& &\mathbb{E}_{\rmA,\{y_i\},\{\rvf_i^{(0)}\}|d_i,y_i}\Bigg(\sum_{j \in \mathcal{N}_i}\frac{\mathbf{f}_j^{(l)}\sqrt{d_i+1}}{\sqrt{d_j+1}}|d_i, y_i\Bigg) 
\\ &=&\mathbb{E}_{\rmA,\{y_i\},\{\rvf_i^{(0)}\}|d_i,y_i, \gN_i}\Bigg(\sum_{j \in \mathcal{N}_i}\frac{\mathbf{f}_j^{(l)}\sqrt{d_i+1}}{\sqrt{d_j+1}}|d_i, y_i, \gN_i\Bigg) \\&=&
d_i\mathbb{E}_{\rmA,\{y_i\},\{\rvf_i^{(0)}\}|d_i,y_i, v_j \in \gN_i}\Bigg(\frac{\mathbf{f}_j^{(l)}\sqrt{d_i+1}}{\sqrt{d_j+1}}|d_i, \xi_{i}^l, v_j \in \gN_i\Bigg)
\label{eq:49}
\end{eqnarray}}

{We note that $v_j$ in Equation~\ref{eq:49} can be any node except $v_i$ due to the equivalence of those nodes.}

Combining Equation~\ref{eq:44} and Equation~\ref{eq:49}, we can obtain:
\begin{equation}
\begin{split}
&\mathbb{E}_{\rmA,\{y_i\},\{\rvf_i^{(0)}\}|d_i,y_i}\Bigg(\mathbf{f}_i^{(l+1)}|d_i,y_i\Bigg) \\=& 
\frac{\xi_i^l\E(\rvf^{(0)}_i|y_i)}{d_i+1} +
\frac{d_i}{d_i+1}\mathbb{E}_{\rmA,\{y_i\},\{\rvf_i^{(0)}\}|d_i,y_i, v_j \in \gN_i}\Bigg(\frac{\mathbf{f}_j^{(l)}\sqrt{d_i+1}}{\sqrt{d_j+1}}|d_i, y_i, v_j \in \gN_i \Bigg)\\=& \frac{\xi_i^l\E(\rvf^{(0)}_i|y_i)}{d_i+1} + \frac{d_i}{d_i+1}\Bigg(\mathbb{E}_{\rmA,\{y_i\},\{\rvf_i^{(0)}\}|d_i,y_i,v_j \in \hat{\mathcal{N}_i^s}, v_j \in \gN_i}\Bigg(\frac{\mathbf{f}_j^{(l)}\sqrt{d_i+1}}{\sqrt{d_j+1}}|d_i, y_i, v_j \in \hat{\mathcal{N}_i^s}, v_j \in \gN_i\Bigg) \\
\cdot& \sP\Bigg(v_j \in \hat{\mathcal{N}_i^s}|v_j \in \gN_i, d_i, y_i\Bigg)
+ \mathbb{E}_{\rmA,\{y_i\},\{\rvf_i^{(0)}\}|d_i,y_i,v_j \not\in \hat{\mathcal{N}_i^s}, v_i \in \gN_i}\Bigg(\frac{\mathbf{f}_j^{(l)}\sqrt{d_i+1}}{\sqrt{d_j+1}}|d_i, y_i, v_j \not\in \hat{\mathcal{N}_i^s}, v_j \in \gN_i\Bigg)
\\ \cdot& \sP\Bigg(v_j \not\in \hat{\mathcal{N}_i^s}, |d_i, y_i, v_j \in \gN_i\Bigg)\Bigg)\\ =& \frac{\xi_i^l\E(\rvf^{(0)}_i|y_i)}{d_i+1} + \Bigg(\frac{d_i\sP(v_j \in \hat{\mathcal{N}_i^s}|d_i, y_i, v_j \in \gN_i)}{{d_i+1} }{\xi_{i}^{l}}'\E(\rvf^{(0)}_i|y_i)- \frac{d_i(1-\sP(v_j \in \hat{\mathcal{N}_i^s}|d_i, y_i, v_j \in \gN_i))}{ {d_i+1}}\rho_{i}^{l}{\xi_{i}^{l}}'\E(\rvf^{(0)}_i|y_i)\Bigg)\\ =& \frac{\xi_{i}^l\E(\rvf^{(0)}_i|y_i)}{d_i+1} + \frac{d_i(\sP(v_j \in \hat{\mathcal{N}_i^s}|d_i, y_i, v_j \in \gN_i)(1+\rho_{i}^l)-\rho_{i}^l){\xi_{i}^{l}}'\E(\rvf^{(0)}_i|y_i)}{d_i+1}
\end{split}
\label{eq:58}
\end{equation}

{Define $\bar{r_i}^{l}\equiv \frac{{\xi_{i}^{l}}'}{{\xi_{i}^{l}}}$. If we know the degree $d_i$, $y_i$, and that a neighbor $v_j$ is in the group $\hat{\mathcal{N}_i^s}$, $\bar{r_i}^{l}$ actually represents the ratio of expected $\rvf_j^{l}r_{i,j}$ to expected $\rvf_i^{l}$. We regard it as the effective related degree of node $v_i$ at the $l$-th
layer. For the initial layer, if $y_j=y_i$, the ratio of  $\rvf_j^{0}r_{i,j}$ to $\rvf_i^{0}$ is $r_{ij}$.
 Recall that the related degree $\ratioi$ at the initial layer is the expected average of $r_{ij}$ in the neighborhood. Thus $\bar{r_i}^{l}$ is an extension of $\ratioi$.
Moreover, let $\hat{h_i^{l}}=\sP(v_j \in \hat{\mathcal{N}_i^s}|v_j \in \gN_i, d_i, y_i)$ represents the probability of a neighbor whose representation has a positive contribution in expectation. This naturally extends the meaning of homophily in deeper layers. Thus, we regard it as the effective homophily of node $v_i$ at $l$-th layer. Given the degree $d_i$ and $y_i$, $\rho_{i}^l$ represents the ratio of the probability that a neighbor will have a positive  rather than a negative contribution to $v_i$. This naturally extends the meaning of $\rho$.}

We further write Equation~\ref{eq:58} as:
$$\frac{\xi_{i}^l}{d_i+1}\Bigg(1+\Bigg((\hat{h_i^{l}}(1+\rho_{i}^l)-\rho_{i}^l)d_i\bar{r_i}^{l})\Bigg)\rvmu$$ and it will have three cases similar to Theorem~\ref{theorem: initial_stage}.


\end{proof}
\subsection{Proof of Theorem~\ref{theorem: signed_messages}}
\label{app:proof3}
\begin{proof}
Similar to Equation~\ref{eq:2}, we can express the conditional expectation of $\rvf_i^{(1)}$ as:
\begin{equation}
\begin{split}
& \E_{\rmA, \{y_i\}, \{\rvf_i^{(0)}\}|d_i, v_i \in \gV_1}\left(\rvf_i^{(1)}|v_i \in \gV_1, d_i\right) \\=& \E_{m_i^0|v_i \in \gV_1, d_i}\left(\E_{\rmA, \{y_i\}, \{\rvf_i^{(0)}\}|d_i, v_i \in \gV_1, m_i^0}\left(\rvf_i^{(1)}|v_i \in \gV_1, d_i, m_i^0\right)\right)  \\
=& \E_{m_i^0|v_i \in \gV_1, d_i}\left(\E_{\rmA, \{y_i\}, \{\rvf_i^{(0)}\}|v_i \in \gV_1, d_i, m_i^0}\Bigg(\frac{\rvf_i^{(0)}}{d_i+1}|v_i \in \gV_1, d_i, m_i^0\right) \\+& \E_{\rmA, \{y_i\}, \{\rvf_i^{(0)}\}|v_i \in \gV_1, d_i, m_i^0}\left(\sum_{j \in \gN_i}\frac{\rvf_j^{(0)}}{\sqrt{d_i+1} \cdot \sqrt{d_j+1}}|v_i \in \gV_1, d_i, m_i^0\right)\Bigg) \\
=& \frac{\rvmu}{d_i+1}\\+& \E_{m_i^0|v_i \in \gV_1, d_i}\Bigg(\E_{\rmA|d_i, v_i \in \gV_1, m_i^0}\Bigg(\sum_{j \in \gN_i}\Bigg(\E_{\{y_i\}, \{\rvf_i^{(0)}\}|\rmA,v_i,v_j \in \gV_1, m_i^0}\left(\frac{\rvf_j^{(0)}}{\sqrt{d_i+1} \cdot \sqrt{d_j+1}}|\rmA, m_i^0, v_j \in \gV_1, v_i \in \gV_1 \right)\\ \cdot& \sP(v_j \in \gV_1|m_i^0, v_i \in \gV_1, \rmA) \\
+& \E_{\{y_i\}, \{\rvf_i^{(0)}\}|\rmA,v_i \in \gV_1, v_j \in \gV_2, m_i^0}\left(\frac{\rvf_j^{(0)}}{\sqrt{d_i+1} \cdot \sqrt{d_j+1}}|\rmA, m_i^0, v_i \in \gV_1, v_j \in \gV_2\right)\cdot \sP(v_j \in \gV_2|m_i^0)\Bigg)\Bigg)\Bigg).
\end{split}
\label{eq:9}  
\end{equation}

Next, we will show how to compute the conditional expectation and conditional probability in the summand.
\begin{equation}
\begin{split}
&\E_{\{y_i\}, \{\rvf_i^{(0)}\}|\rmA,v_i,v_j \in \gV_1, m_i^0}\left(\frac{\rvf_j^{(0)}}{\sqrt{d_i+1} \cdot \sqrt{d_j+1}}|m_i^0, v_i, v_j \in \gV_1,\rmA\right) \\=& \E_{\{y_i\}, \{\rvf_i^{(0)}\}|\rmA,v_i \in \gV_1, m_i^0, v_j}\left(\frac{\rvf_j^{(0)}}{\sqrt{d_i+1} \cdot \sqrt{d_j+1}}|\rmA, m_i^0, v_i \in \gV_1, v_j \in \gV_1, v_j \text{wrongly send information}\right) \\
& \cdot \sP(v_j \text{wrongly send information}|\rmA, m_i^0, v_i \in \gV_1, v_j \in \gV_1) \\
+& \E_{\{y_i\}, \{\rvf_i^{(0)}\}|\rmA,v_i \in \gV_1, m_i^0, v_j}\left(\frac{\rvf_j^{(0)}}{\sqrt{d_i+1} \cdot \sqrt{d_j+1}}|\rmA, m_i^0, v_i \in \gV_1, v_j \in \gV_1, v_j \text{correctly send information}\right) \\
& \cdot \sP(v_j \text{correctly send information}|\rmA, m_i^0, v_i \in \gV_1, v_j \in \gV_1).
\end{split}
\label{eq:10}
\end{equation}
Combined with the independence assumption, we have:
\begin{equation}
\begin{split}
\sP(v_j \text{wrongly send information}|\rmA, m_i^0, v_i \in \gV_1, v_j \in \gV_1) =
%
%
m_i^0.
\end{split}
\label{eq:11}
\end{equation}
Similarly, we obtain:
\begin{equation}
\sP(v_j \text{correctly send information}|\rmA, m_i^0, v_i \in \gV_1, v_j \in \gV_1) = 1-m_i^0.
\label{eq:12}
\end{equation}
Then Equation~(\ref{eq:10}) can be rewritten as:
\begin{equation}
\begin{split}
&\E_{\{y_i\}, \{\rvf_i^{(0)}\}|\rmA,v_i,v_j \in \gV_1, m_i^0}\left(\frac{\rvf_j^{(0)}}{\sqrt{d_i+1} \cdot \sqrt{d_j+1}}|m_i^0, v_i, v_j \in \gV_1,\rmA\right) \\ =& \E_{\{y_i\}, \{\rvf_i^{(0)}\}|\rmA,v_i \in \gV_1, m_i^0, v_j}\left(\frac{\rvf_j^{(0)}}{\sqrt{d_i+1} \cdot \sqrt{d_j+1}}|\rmA, m_i^0, v_i \in \gV_1, v_j \in \gV_1, v_j \text{wrongly send information}\right)\cdot m_i^0 \\
+& \E_{\{y_i\}, \{\rvf_i^{(0)}\}|\rmA,v_i \in \gV_1, m_i^0, v_j}\left(\frac{\rvf_j^{(0)}}{\sqrt{d_i+1} \cdot \sqrt{d_j+1}}|\rmA, m_i^0, v_i \in \gV_1, v_j \in \gV_1, v_j \text{correctly send information}\right)\cdot (1-m_i^0) \\
=& -\frac{m_i^0\rvmu}{\sqrt{d_i+1} \cdot \sqrt{d_j+1}} + \frac{(1-m_i^0)\rvmu}{\sqrt{d_i+1} \cdot \sqrt{d_j+1}} \\
=&
\frac{(1-2m_i^0)\rvmu}{\sqrt{d_i+1} \cdot \sqrt{d_j+1}}.
\end{split}
\label{eq:13}
\end{equation}
\vspace{-0.3cm}
Similarly, we have:
\vspace{-0.2cm}
\begin{equation}
\E_{\{y_i\}, \{\rvf_i^{(0)}\}|\rmA,v_i \in \gV_1, v_j \in \gV_2, m_i^0 }\left(\frac{\rvf_j^{(0)}}{\sqrt{d_i+1} \cdot \sqrt{d_j+1}}|\rmA, m_i^0, v_i \in \gV_1, v_j \in \gV_2\right) = \frac{(1-2m_i^0)\rho\rvmu}{\sqrt{d_i+1}. \cdot \sqrt{d_j+1}}.
\label{eq:14}
\end{equation}
Consider the independence between $m_i^0$ and node class \& degrees, and insert Equation~(\ref{eq:13}) and Equation~(\ref{eq:14}) into Equation~(\ref{eq:9}), we will have:
\begin{equation}
\begin{split}
& \E_{\rmA, \{y_i\}, \{\rvf_i^{(0)}\}|d_i, v_i \in \gV_1}\left(\rvf_i^{(1)}|v_i \in \gV_1, d_i\right) \\ =& \frac{\rvmu}{d_i+1} + \E_{m_i^0|v_i \in \gV_1, d_i}\Bigg(\E_{\rmA|d_i, v_i \in \gV_1, m_i^0}\left(\sum_{j \in \gN_i}(\frac{(1-2m_i^0)k_i\rvmu}{\sqrt{d_i+1} \cdot \sqrt{d_j+1}}+\frac{(1-2m_i^0)(1-k_i)\rho\rvmu}{\sqrt{d_i+1} \cdot \sqrt{d_j+1}})\right)\Bigg) \\
=& \frac{\rvmu}{d_i+1} + \E_{m_i^0|v_i \in \gV_1, d_i}\Bigg(\E_{\rmA|d_i, v_i \in \gV_1, m_i^0}\left( \sum_{j \in \gN_i}\frac{(1-2m_i^0)(\rho+(1-\rho)k_i)\rvmu}{\sqrt{d_i+1} \cdot \sqrt{d_j+1}}\right)\Bigg) \\
%
%
=& \frac{\rvmu}{d_i+1} + \E_{m_i^0|v_i \in \gV_1, d_i}\Bigg(\left(\frac{(1 - 2m_i^0)(\rho+(1-\rho)h_i)d_i\ratioi}{d_i+1}\right)\rvmu\Bigg)\\
=& \left(\frac{1 + (1 - 2e_i^0)(\rho+(1-\rho)h_i)d_i\ratioi}{d_i+1}\right)\rvmu \equiv \gamma_i^1 \E(\rvf^{(0)}_i).
\end{split}
\label{eq:15}
\end{equation}
When $h_i \leq 1$, $\rho+(1-\rho)h_i=(1-h_i)\rho+h_i>0.$ Define $\epsilon' \equiv (1 - 2e_i^0)(\rho+(1-\rho)h_i)$. To obtain the ranges of $\gamma_i^1$, when $e_i^0 \geq 0.5$, $\epsilon' \leq 0$, it resembles the derivations of CASE 1 in Proof~\ref{app:proof1}; when $e_i^0 < 0.5$, $0 < \epsilon \ratioi \leq 1$, it resembles the derivations of CASE 2; when $e_i^0 < 0.5$, $\epsilon \ratioi > 1$, it resembles the derivations of CASE 3.
\end{proof}

\subsection{Movements \& Misclassification Rate}
\label{app:mr}
In this section, we illustrate why the movements of node representations are a good indicator of SGC's performance. The misclassification rate of an $L$-layer SGC can be studied through its last logistic-regression layer, the input to which is $\{\rvf_i^{(L)}\}$. To study the misclassification rate of an $(L+1)$-layer SGC, we can view the input to the last layer  $\{\rvf_i^{(L+1)}\}$ as being moved from $\{\rvf_i^{(L)}\}$. The misclassification rate is closely related to the decision boundary and will be our tool for studying the change of the misclassification rate. Our goal is to study how the movements change the decision boundary, which in turn affect the misclassification rate. Recall that we are studying the SGC model for a binary classification task.

\subsubsection{Preliminaries}
\label{app:lemma_1}
\textbf{Lemma} In SGC, the decision boundaries w.r.t. $\{\rvf_i^{(L)}\}$ in multi-class classification are linear hyperplanes. In particular, the decision boundary is a single hyperplane for binary classification.

\begin{proof}
{The loss function for SGC is: \texttt{CrossEntropy}(\texttt{Softmax}($\rmF^{(L)}\rmW^{(L)}+\rvb^{(L)}$), $y_i$). Rewrite $\rmW^{(L)}$ into an array of column vectors: $\rmW^{(L)}=[\rvw_0^{(L)}, \rvw_1^{(L)}, \dots \rvw_{|\gL|}^{(L)}]$, $\rvb^{(L)}$ into an array of scalars: $\rvb^{(L)}=[b_0^{(L)}, b_1^{(L)}, \dots b_{|\gL|}^{(L)}]$, and $\rmF^{(L)}$ into an array of row vectors: $\rmF^{(L)}=[\rvf_0^{(L)}; \rvf_1^{(L)}; \dots \rvf_{|\gV|}^{(L)}]$ ($[\cdot ; \cdot]$ means stacking vertically). Let $p_{i,c} \equiv \frac{e^{\rvf_i^{(L)}\rvw_c^{(L)}+b_c^{(L)}}}{\sum_j e^{\rvf_i^{(L)}\rvw_j^{(L)}+b_j^{(L)}}}$ and let $y_{i,c} \equiv \begin{cases} 1, & \text{iff } y_i=c \\ 0, & \text{otherwise} \end{cases}$
then we rewrite the loss function as:
\begin{equation}
    -\frac{1}{n}\sum_{i=1}^n{\sum_{c=1}^{|\gL|}y_{i,c}\cdot\text{log}(p_{i,c})},
\end{equation}
$n$ is the number of nodes used for training. To predict that the node $v_i$ belongs to class $c$, we require: $\forall c'\neq c$, $p_{i,c} > p_{i,c'}$. Thus, we have: 
\begin{equation}
    \frac{e^{\rvf_i^{(L)}\rvw_c^{(L)}+b_c^{(L)}}}{\sum_j e^{\rvf_i^{(L)}\rvw_j^{(L)}+b_j^{(L)}}}>\frac{e^{\rvf_i^{(L)}\rvw_{c'}^{(L)}+b_{c'}^{(L)}}}{\sum_j e^{\rvf_i^{(L)}\rvw_{j}^{(L)}+b_j^{(L)}}}.
\end{equation} 
It is equivalent to: 
\begin{equation}
    {\rvf_i^{(L)}\rvw_c^{(L)}+b_c^{(L)}}>\rvf_i^{(L)}\rvw_{c'}^{(L)}+b_{c'}^{(L)}.
\end{equation} 
It shows that the decision region of class $c$ is decided by a set of hyper-planes: 
\begin{equation}
    \rvx(\rvw_c^{(L)}-\rvw_{c'}^{(L)})+(b_c^{(L)}-b_{c'}^{(L)})>0.
\end{equation}}
\end{proof}
{$\rvx$ is the vector variable that describes the plane. Any vectors that satisfy the equation are on this plane. We note that for binary classification, there is one hyper-plane.}

The decision boundary of an $(L+1)$-layer SGC can be viewed as a perturbation to the decision boundary of an $L$-layer SGC due to the movements from $\{\rvf_i^{(L)}\}$ to $\{\rvf_i^{(L+1)}\}$. Due to the Lemma, we can use the hyperplane $\rvx\rvw+b=0$ to represent the decision boundary of $\{\rvf_i^{(L)}\}$, 
where $\rvw$ is a unit vector, $b$ is a scalar, and the $\{\rvf_i^{(L)}\}$ are classified to class 1 if $\rvf_i^{(L)}\rvw+b>0$.
Suppose when $\rvw=\rvw^*$ and $b=b^*$, we find an optimal hyperplane $\rvx{\rvw^*}+b^*=0$ that achieves the lowest total misclassification rate (\S~\ref{section: background}) for $\{\rvf_i^{(L)}\}$. Due to the movements of node representations, the {new optimal hyperplane is decided by}: $\rvx{\rvw^*}'+{b^*}'=0$.
In the following theorem, we show that, \ul{under constraints, moving towards the original decision boundary with a non-zero step results in a non-decreasing total misclassification rate under the new decision boundary.}

\subsubsection{Theorem}
\label{app:lemma_2}
\textbf{Theorem} Moving representations $\{\rvf_i^{(L)}\}$ from class 1 by adding $-t{\rvw^*}^T$, $t>0$, s.t. ${\rvw^*}^T{\rvw^*}'>0$, the new total misclassification rate is no less than the misclassification rate before the movements.

{This Theorem studies the movements that bring the node representations closer to the original decision boundary; we will prove that moving towards the original decision boundary by a non-zero step is not beneficial (i.e., harmful) to the SGC's performance.}

\begin{proof}
We prove it by contradiction and suppose that after the movements, the total misclassification rate is lowered. 
{We denote the conditional distribution of $\{\rvf_i^{(L)}\}$ as $f_1^{L}(\rvx)$ conditioned on that they are from class 1 and $f_2^{L}(\rvx)$ conditioned on that they are from class 1 or class 2, respectively.} 
To note that, the distributions of different layers are different. 

{After the movements, the representations from class 1 become: $\{\rvf_i^{(L)}-t{\rvw^*}^T\}, t>0$. {$t{\rvw^*}^T$ represents moving towards the original decision boundary along the norm direction by a non-zero step.} This causes a corresponding change in the conditional PDF given that they are from class 1: $f_1^L(\rvx+t{\rvw^*}^T)$. 

{Recall that in \S~\ref{section: analysis}, the parameters for the original optimal hyperplane ${\rvw^*}$ and ${b^*}$, and the parameters for the later optimal hyperplane ${\rvw^*}'$ and ${b^*}'$ are normalized such that when $\rvx{\rvw^*}+{b^*}>0$ and $\rvx{\rvw^*}'+{b^*}'>0$, we predict class 1.} 

The new misclassification rate conditioned on class 1 is: 
\begin{equation}
   M_t'=\int_{\rvx{\rvw^*}'+{b^*}'<0}{f_1^L(\rvx+t{\rvw^*}^T)d\rvx}=\int_{(\rvx-t{\rvw^*}^T){\rvw^*}'+{b^*}'<0}{f_1^L(\rvx)d\rvx}. 
\end{equation}}
{The new misclassification rate conditioned on class 2 is: 
\begin{equation}
    M_r'=\int_{\rvx{\rvw^*}'+{b^*}'>0}{f_2^L(\rvx)d\rvx}.
\end{equation}
The new total misclassification rate is:
{
\begin{equation}
\label{lemma2:total}
\begin{split}
    M' &= M_t'\sP(v_i \in \gV_1)+M_r'\sP(v_i \in \gV_2)\\ &= M_t'\frac{\sP(v_i \in \gV_1)}{\sP(v_i \in \gV_1)+\sP(v_i \in \gV_2)}+M_r'\frac{\sP(v_i \in \gV_2)}{\sP(v_i \in \gV_1)+\sP(v_i \in \gV_2)}\\ &=\frac{\rho}{\rho+1}M_t'+\frac{1}{\rho+1}M_r'.
\end{split}
\end{equation}}} 
{Next, we will prove that if the total misclassification is lowered, $\rvx\rvw^*+b^*=0$ is not the optimal hyper-plane before the movements which should achieve the lowest total misclassification rate.} 

{Consider a hyper-plane $\rvx{\rvw^*}'+{b^*}'-\frac{t{{\rvw^*}^T}{\rvw^*}'}{2}=0$.}
{Given that ${{\rvw^*}^T}{\rvw^*}'>0$ and $t>0$, for $\forall \rvx,$ s.t. $\rvx{\rvw^*}'+{b^*}'-\frac{t{{\rvw^*}^T}{\rvw^*}'}{2}<0$, we have: $(\rvx-t{\rvw^*}^T){\rvw^*}'+{b^*}'= \rvx{\rvw^*}'+{b^*}'-\frac{t{{\rvw^*}^T}{\rvw^*}'}{2}-\frac{t{{\rvw^*}^T}{\rvw^*}'}{2}<0$. It means: 
\begin{equation}
    \{\rvx|(\rvx-t{\rvw^*}^T){\rvw^*}'+{b^*}'<0\} \supseteq	\{\rvx|\rvx{\rvw^*}'+{b^*}'-\frac{t{{\rvw^*}^T}{\rvw^*}'}{2}<0\}.
\end{equation}}
{Because $f_1^L(\rvx)$ is a PDF which is a nonnegative function, 
\begin{equation}
\label{lemma2:ineq_1}
    \int_{(\rvx-t{\rvw^*}^T){\rvw^*}'+{b^*}'<0}{f_1^L(\rvx)d\rvx}\geq \int_{\rvx{\rvw^*}'+{b^*}'-\frac{t{{\rvw^*}^T}{\rvw^*}'}{2}<0}{f_1^L(\rvx)d\rvx}.
\end{equation}} 
{Similarly, we can obtain
\begin{equation}
    \{\rvx|\rvx{\rvw^*}'+{b^*}'>0\} \supseteq	\{\rvx|\rvx{\rvw^*}'+{b^*}'-\frac{t{{\rvw^*}^T}{\rvw^*}'}{2}>0\}.
\end{equation}
Thus, 
\begin{equation}
\label{lemma2:ineq_2}
    \int_{\rvx{\rvw^*}'+{b^*}'>0}{f_2^L(\rvx)d\rvx} \geq \int_{\rvx{\rvw^*}'+{b^*}'-\frac{t{{\rvw^*}^T}{\rvw^*}'}{2}>0}{f_2^L(\rvx)d\rvx}.
\end{equation} 
Define 
\begin{equation}
    M_{f} \equiv \frac{\rho}{\rho+1}\int_{\rvx{\rvw^*}'+{b^*}'-\frac{t{{\rvw^*}^T}{\rvw^*}'}{2}<0}{f_1^L(\rvx)d\rvx}+\frac{1}{\rho+1}\int_{\rvx{\rvw^*}'+{b^*}'-\frac{t{{\rvw^*}^T}{\rvw^*}'}{2}>0}{f_2^L(\rvx)d\rvx}.
\end{equation} 
The quantity $M_{f}$ represents the total misclassification rate before the movements if the decision boundary is $\rvx{\rvw^*}'+{b^*}'-\frac{t{{\rvw^*}^T}{\rvw^*}'}{2}=0$.
Given Eq.~\ref{lemma2:total}, \ref{lemma2:ineq_1} and \ref{lemma2:ineq_2}, we have: 
\begin{equation}
    M' \geq M_{f}. 
\end{equation} 
Let $M$ denotes the total classification rate before the movements. Based on the assumption that the total misclassification rate is lowered after the movements ($M > M'$), we have:
\begin{equation}
\label{lemma2:conclusion}
    M > M_{f}. 
\end{equation}
Eq.~\ref{lemma2:conclusion} indicates that we find a hyper-plane $\rvx{\rvw^*}'+{b^*}'-\frac{t{{\rvw^*}^T}{\rvw^*}'}{2}=0$, which yields smaller total misclassification rate than $\rvx{\rvw^*}+{b^*}=0$ before the movements. This contradicts to the fact that $\rvx{\rvw^*}+{b^*}=0$ is the optimal hyper-plane before the movements.}
\end{proof}
Note that the special case where the representations of the two classes swap positions (e.g, bipartite graphs) violates the condition ${\rvw^*}^T{\rvw^*}'>0$ and leads to different conclusions. We refer to the condition ${\rvw^*}^T{\rvw^*}'>0$ as "non-swapping condition" and throughout the paper, we analyze SGC under that. The theorem shows that, \ul{under the "non-swapping condition", if $\{\rvf_i^{(l)}\}$ move towards the original decision boundary (or the other class), SGC tends to perform worse.}

\section{Additional Experiments}

\subsection{Ablation study}
\label{subsec:ablation}
\setlength{\tabcolsep}{3pt}
\setlength{\tabcolsep}{3pt}
\begin{table*}[h]
\centering
\vspace{-0.3cm}
\caption{Ablation study: both the structure-based edge correction and the feature-based edge correction can alleviate the oversmoothing and heterophily problems. Structure-based edge correction has consistent benefits over both homophilous and heterophilous datasets while feature-based edge correction has more benefits on heterophilous datasets. Best performance of each model is highlighted in gray.}
\label{tab:ablation}
\begin{adjustbox}{width=1\linewidth}
{\footnotesize
\begin{tabular}{ccccccccccccccc}
\toprule
                                \multicolumn{1}{r}{\textbf{Layers}} 
                                      && \textbf{2}           & \multicolumn{1}{c}{\textbf{4}} & \multicolumn{1}{c}{\textbf{8}} & \multicolumn{1}{c}{\textbf{16}} & \multicolumn{1}{c}{\textbf{32}} & \multicolumn{1}{c}{\textbf{64}} &&  \multicolumn{1}{c}{\textbf{2}}           & \multicolumn{1}{c}{\textbf{4}} & \multicolumn{1}{c}{\textbf{8}} & \multicolumn{1}{c}{\textbf{16}} & \multicolumn{1}{c}{\textbf{32}} & \multicolumn{1}{c}{\textbf{64}}  \\ \cmidrule{1-1} \cmidrule{3-8} \cmidrule{10-15} 
&& \multicolumn{6}{c}{\textbf{Cora} ($h$=0.81)}     & & \multicolumn{6}{c}{\textbf{Citeseer} ($h$=0.74)}                                                                         \\ 
\cline{3-8} \cline{10-15}
                                    
                                        \multicolumn{1}{c}{Base}    &&   
                                        \cellcolor{gray!15}$ 86.56{\scriptstyle\pm1.21}$         & \multicolumn{1}{c}{$ 86.04{\scriptstyle\pm0.72}$}    &        \multicolumn{1}{c}{$ 85.51{\scriptstyle\pm1.51}$}     &       \multicolumn{1}{c}{$ 85.33{\scriptstyle\pm0.72}$}      &     \multicolumn{1}{c}{$ 85.37{\scriptstyle\pm1.58}$}       &    \multicolumn{1}{c}{$ 72.17{\scriptstyle\pm8.89}$}   & &  
                                        \cellcolor{gray!15}$76.51{\scriptstyle\pm1.63}$ & $75.03{\scriptstyle\pm1.67}$ & $73.96{\scriptstyle\pm1.52}$ & $73.59{\scriptstyle\pm1.51}$& $71.91{\scriptstyle\pm1.94}$ & $32.08{\scriptstyle\pm15.74}$                                         \\ 
                                        
                                      \multicolumn{1}{c}{\texttt{+str}}    &&  
                                      \cellcolor{gray!15}$86.72{\scriptstyle\pm1.29}$      &          \multicolumn{1}{c}{$86.02{\scriptstyle\pm0.97}$}           & \multicolumn{1}{c}{$85.49{\scriptstyle\pm1.32}$}           & \multicolumn{1}{c}{$85.27{\scriptstyle\pm1.59}$}            & \multicolumn{1}{c}{$85.27{\scriptstyle\pm1.51}$}            & \multicolumn{1}{c}{$84.21{\scriptstyle\pm1.22}$}   &   &                   
                                      \cellcolor{gray!15}$76.63{\scriptstyle\pm1.38}$  & $74.64{\scriptstyle\pm1.97}$&  $74.15{\scriptstyle\pm1.61}$&  $73.73{\scriptstyle\pm1.31}$&  $73.61{\scriptstyle\pm1.84}$&  $70.56{\scriptstyle\pm2.27}$
                                      \\  \multicolumn{1}{c}{\texttt{+feat}} &&  
                                      $84.81{\scriptstyle\pm1.63}$&   \cellcolor{gray!15}$86.06{\scriptstyle\pm1.7}$&   $85.67{\scriptstyle\pm1.26}$&   $85.39{\scriptstyle\pm0.97}$&   $84.85{\scriptstyle\pm0.98}$&   $78.57{\scriptstyle\pm6.73}$   & &    
                                      \cellcolor{gray!15}$77.13{\scriptstyle\pm1.69}$&  $74.56{\scriptstyle\pm2.02}$&   $73.64{\scriptstyle\pm1.65}$&   $72.31{\scriptstyle\pm2.32}$&   $71.98{\scriptstyle\pm3.44}$&   $68.68{\scriptstyle\pm6.72}$
                        \\  \multicolumn{1}{c}{\texttt{+str,feat}} &&
                        \cellcolor{gray!15}$86.96{\scriptstyle\pm1.38}$ & $86.20{\scriptstyle\pm0.89}$ & $85.63{\scriptstyle\pm0.78}$ &
                              $85.47{\scriptstyle\pm1.18}$ & $85.55{\scriptstyle\pm1.66}$ & $77.81{\scriptstyle\pm7.95}$     &  & 
                              \cellcolor{gray!15}$76.81{\scriptstyle\pm1.71}$ & $74.68{\scriptstyle\pm1.97}$ & $74.69{\scriptstyle\pm2.35}$ &
                              $73.28{\scriptstyle\pm1.45}$ & $71.81{\scriptstyle\pm2.28}$ & $69.91{\scriptstyle\pm3.97}$ 
                              \\ \midrule 
&& \multicolumn{6}{c}{\textbf{Cornell ($h$=0.3)}} && \multicolumn{6}{c}{\textbf{Chameleon ($h$=0.23)}}                                                                                                                                                                                                                               \\ 
\cline{3-8} \cline{10-15} 
                                      \multicolumn{1}{c}{Base}    && \cellcolor{gray!15}$61.89{\scriptstyle\pm3.72}$ & $60.00{\scriptstyle\pm5.24}$ & $58.92{\scriptstyle\pm5.24}$ &
                              $56.49{\scriptstyle\pm5.73}$ & $58.92{\scriptstyle\pm3.15}$ & $49.19{\scriptstyle\pm16.70}$          &&
\cellcolor{gray!15}$64.98{\scriptstyle\pm1.84}$ & $62.65{\scriptstyle\pm3.09}$ & $62.43{\scriptstyle\pm3.28}$ & $54.69{\scriptstyle\pm2.58}$& $47.68{\scriptstyle\pm2.63}$ & $29.74{\scriptstyle\pm5.21}$
\\  
                                      \multicolumn{1}{c}{\texttt{+str}}    && \cellcolor{gray!15}$63.78{\scriptstyle\pm5.57}$ & $62.70{\scriptstyle\pm5.90}$ & $59.46{\scriptstyle\pm4.52}$ &
                              $56.49{\scriptstyle\pm5.73}$ & $57.57{\scriptstyle\pm4.20}$ & $58.92{\scriptstyle\pm3.15}$    &&
$66.54{\scriptstyle\pm2.19}$ & $68.31{\scriptstyle\pm2.70}$ & \cellcolor{gray!15}$68.99{\scriptstyle\pm2.38}$ & $67.68{\scriptstyle\pm3.70}$& $56.86{\scriptstyle\pm8.80}$ & $41.95{\scriptstyle\pm9.56}$
\\  
                                      \multicolumn{1}{c}{\texttt{+feat}} && \cellcolor{gray!15}$85.41{\scriptstyle\pm7.27}$&  $76.76{\scriptstyle\pm7.07}$& $70.00{\scriptstyle\pm5.19}$&  $67.57{\scriptstyle\pm9.44}$&  $63.24{\scriptstyle\pm6.07}$&  $63.24{\scriptstyle\pm6.53}$    && 
\cellcolor{gray!15}$65.31{\scriptstyle\pm3.20}$&  $53.55{\scriptstyle\pm6.35}$& $53.05{\scriptstyle\pm2.28}$&  $51.93{\scriptstyle\pm4.00}$&  $57.17{\scriptstyle\pm3.39}$&  $51.93{\scriptstyle\pm8.95}$
\\  
                                      \multicolumn{1}{c}{\texttt{+str,feat}}            && \cellcolor{gray!15}$84.32{\scriptstyle\pm6.37}$&  $78.92{\scriptstyle\pm8.09}$& $73.51{\scriptstyle\pm5.90}$&  $70.81{\scriptstyle\pm5.64}$&  $68.11{\scriptstyle\pm5.14}$&  $62.43{\scriptstyle\pm6.67}$ &&
\cellcolor{gray!15}$65.75{\scriptstyle\pm1.81}$ & $61.49{\scriptstyle\pm7.38}$ & $53.73{\scriptstyle\pm7.79}$ &
                               $52.43{\scriptstyle\pm5.37}$ & $55.92{\scriptstyle\pm5.14}$ & $56.95{\scriptstyle\pm3.93}$ 

                              \\ \bottomrule

\end{tabular}
}
\end{adjustbox}
\end{table*}
We now study the impact of our proposed mechanisms (structure- and feature-based edge correction, Section~\ref{section: model_design}). To better show their effects, we add each design choice to a base model and track the changes of the performance in the node classification task. We choose a GCN~\citep{kipf2016semi} variant as the base model, which uses weight bias, residual connections (\textit{instead of decaying aggregation}), and \texttt{Elu} as its non-linearity. We use this variant because it is more robust to oversmoothing than the vanilla GCN model~\citep{kipf2016semi}.
We denote the model that incorporates the structure-based edge correction as \texttt{+str}, and the model that incorporates feature-based edge correction as \texttt{+feat}. The model that uses both designs is denoted as \texttt{+str,feat}. 
Table~\ref{tab:ablation} gives the accuracy of the models in the semi-supervised node classification task for different layers. 

We observe that both mechanisms alleviate the oversmoothing problem. Specifically, the base model has a sharp performance decrease after 32 layers, while the other models have significantly higher performance. 
In general, the \texttt{+str} model is better than \texttt{+feat} in alleviating oversmoothing, and its performance decreases the least  at the 64-th layer on the Cora, Citeseer and Cornell datasets. We also observe that the \texttt{+str} model achieves a 5\% increase in accuracy at the 8-th layer on the Chameleon dataset, which can possibly explain the large performance gain of \method in Table~\ref{tab:real-results}.

From Table~\ref{tab:ablation}, we can see that both mechanisms improve the performance of the base model under the heterophily settings, and the \texttt{+feat} model has an advantage over the \texttt{+str} model on heterophilous datasets. For instance, on the Cornell dataset, the \texttt{+feat} model achieves around 24\% and 22\% gain over the base model and the \texttt{+str} model, respectively. However, on homophilous datasets, the benefits from signed messages are  limited. 
{This is because when the effective homophily $\heterophilyil$ is high and error rate $e_i^0$ is low, $\lim_{\heterophilyil \to 1, e_i^0 \to 0} \frac{(1+\rho)\heterophilyil-\rho}{(1 - 2e_i^0)(\rho+(1-\rho)\heterophilyil)}=1$, and the number of nodes in case 3 is similar to when we do not use signed edges. As we know, the benefits of the model come from nodes in case 3, so it explains why signed edges have limited benefits on homophilous graphs.}

\subsection{Batch norm \& Layer norm}
\label{app:batchnorm_layernorm}
In Section~\ref{section: model_design}, we use decaying aggregation instead of other normalizing mechanisms. Other mechanisms, such as batch or layer norm, may be seen as solutions to the heterophily and oversmoothing problems. 
However, batch norm cannot compensate for the dispersion of the node representations (Section~\ref{section: analysis}) due to different degrees and homophily levels of the nodes. Although, to some extent, it reduces the speed by which the representations of the susceptible nodes (case 1 \& 2) move towards the other class (good for oversmoothing), it also prevents the representations of the nodes that could benefit from the propagation (case 3) from increasing the distances (drop in accuracy). 
Layer norm is better at overcoming the dispersion effect but may lead to a significant accuracy drop in some datsets when a subset of features are more important than the others. Thus, we \textbf{do not} use any of these normalizations. 
Next, we provide experiments to show the effects of batch norm and layer norm. We use the following base model (the same model used in Section \ref{subsec:ablation}): a GCN~\cite{kipf2016semi} with weight bias, \texttt{Elu} non-linearity and residual connection. We do \textbf{not} include any of our designs so as to exclude any other factors that can affect the performance. The models we compare against are \texttt{+BN} and \texttt{+LN}, which 
represent the models that add batch norm and layer norm right before the non-linear activation, respectively. 

\setlength{\tabcolsep}{3pt}
\begin{table*}[h]
\vspace{-0.2cm}
\centering
\caption{Effects of using batch norm \& layer norm: decrease in accuracy but improvement in oversmoothing. Best performance of each model across different layers is highlighted in gray.}
\label{tab:bn_ln}
\vspace{-0.2cm}
\begin{adjustbox}{width=\linewidth}
{\footnotesize
\begin{tabular}{ccccccccccccccc}
\toprule
                                \multicolumn{1}{r}{\textbf{Layers}} 
                                      && \textbf{2}           & \multicolumn{1}{c}{\textbf{4}} & \multicolumn{1}{c}{\textbf{8}} & \multicolumn{1}{c}{\textbf{16}} & \multicolumn{1}{c}{\textbf{32}} & \multicolumn{1}{c}{\textbf{64}} &&  \multicolumn{1}{c}{\textbf{2}}           & \multicolumn{1}{c}{\textbf{4}} & \multicolumn{1}{c}{\textbf{8}} & \multicolumn{1}{c}{\textbf{16}} & \multicolumn{1}{c}{\textbf{32}} & \multicolumn{1}{c}{\textbf{64}}  \\ \cmidrule{1-1} \cmidrule{3-8} \cmidrule{10-15} 
&& \multicolumn{6}{c}{\textbf{Cora} ($h$=0.81)}     & & \multicolumn{6}{c}{\textbf{Citeseer} ($h$=0.74)}                                                                         \\ 
\cline{3-8} \cline{10-15}
                                    
                                        \multicolumn{1}{c}{Base}    &&   
                                        \cellcolor{gray!15}$ 86.56{\scriptstyle\pm1.21}$         & \multicolumn{1}{c}{$ 86.04{\scriptstyle\pm0.72}$}    &        \multicolumn{1}{c}{$ 85.51{\scriptstyle\pm1.51}$}     &       \multicolumn{1}{c}{$ 85.33{\scriptstyle\pm0.72}$}      &     \multicolumn{1}{c}{$ 85.37{\scriptstyle\pm1.58}$}       &    \multicolumn{1}{c}{$ 72.17{\scriptstyle\pm8.89}$}   & &  
                                        \cellcolor{gray!15}$76.51{\scriptstyle\pm1.63}$ & $75.03{\scriptstyle\pm1.67}$ & $73.96{\scriptstyle\pm1.52}$ & $73.59{\scriptstyle\pm1.51}$& $71.91{\scriptstyle\pm1.94}$ & $32.08{\scriptstyle\pm15.74}$                                         \\ 
                                        
                                      \multicolumn{1}{c}{\texttt{+BN}}    &&  
                                      $84.73{\scriptstyle\pm1.10}$      &          \multicolumn{1}{c}{$83.76{\scriptstyle\pm1.61}$}           & \multicolumn{1}{c}{$83.94{\scriptstyle\pm1.51}$}           & \multicolumn{1}{c}{$84.57{\scriptstyle\pm1.22}$}            & \multicolumn{1}{c}{$84.63{\scriptstyle\pm1.58}$}            & \cellcolor{gray!15}{$85.17{\scriptstyle\pm1.18}$}   &   &                   
                                      $71.62{\scriptstyle\pm1.48}$  & $71.58{\scriptstyle\pm1.00}$&  $72.18{\scriptstyle\pm1.39}$&  $72.45{\scriptstyle\pm1.42}$&  \cellcolor{gray!15}$72.76{\scriptstyle\pm1.31}$&  $72.61{\scriptstyle\pm1.41}$
                                      \\  \multicolumn{1}{c}{\texttt{+LN}} &&  
                                      $84.73{\scriptstyle\pm1.63}$&   $86.60{\scriptstyle\pm1.01}$&   \cellcolor{gray!15}$86.72{\scriptstyle\pm1.36}$&   $86.08{\scriptstyle\pm1.16}$&   $85.67{\scriptstyle\pm1.23}$&   $85.13{\scriptstyle\pm1.20}$   & &    
                                      \cellcolor{gray!15}$76.11{\scriptstyle\pm1.80}$&  $74.02{\scriptstyle\pm2.77}$&   $75.00{\scriptstyle\pm1.95}$&   $74.50{\scriptstyle\pm0.96}$&   $74.49{\scriptstyle\pm2.10}$&   $73.94{\scriptstyle\pm2.03}$
                        \\   \midrule 
&& \multicolumn{6}{c}{\textbf{Cornell ($h$=0.3)}} && \multicolumn{6}{c}{\textbf{Chameleon ($h$=0.23)}}                                                                                                                                                                                                                               \\ 
\cline{3-8} \cline{10-15} 
                                      \multicolumn{1}{c}{Base}    && \cellcolor{gray!15}$61.89{\scriptstyle\pm3.72}$ & $60.00{\scriptstyle\pm5.24}$ & $58.92{\scriptstyle\pm5.24}$ &
                              $56.49{\scriptstyle\pm5.73}$ & $58.92{\scriptstyle\pm3.15}$ & $49.19{\scriptstyle\pm16.70}$          &&
\cellcolor{gray!15}$64.98{\scriptstyle\pm1.84}$ & $62.65{\scriptstyle\pm3.09}$ & $62.43{\scriptstyle\pm3.28}$ & $54.69{\scriptstyle\pm2.58}$& $47.68{\scriptstyle\pm2.63}$ & $29.74{\scriptstyle\pm5.21}$
\\  
                                      \multicolumn{1}{c}{\texttt{+BN}}    && $58.38{\scriptstyle\pm6.42}$ & \cellcolor{gray!15}$59.19{\scriptstyle\pm4.59}$ & $55.41{\scriptstyle\pm6.65}$ &
                              $57.30{\scriptstyle\pm3.15}$ & $57.57{\scriptstyle\pm6.29}$ & $57.02{\scriptstyle\pm6.19}$    &&
$60.88{\scriptstyle\pm2.24}$ & $61.38{\scriptstyle\pm2.17}$ & $61.84{\scriptstyle\pm4.08}$ & \cellcolor{gray!15}$61.97{\scriptstyle\pm3.01}$& $59.04{\scriptstyle\pm3.79}$ & $57.84{\scriptstyle\pm3.67}$
\\  
                                      \multicolumn{1}{c}{\texttt{+LN}} && $58.11{\scriptstyle\pm6.19}$&  $55.68{\scriptstyle\pm6.19}$& $58.92{\scriptstyle\pm7.63}$&  \cellcolor{gray!15}$59.19{\scriptstyle\pm3.07}$&  $58.92{\scriptstyle\pm3.15}$&  $58.00{\scriptstyle\pm3.03}$    && 
$61.86{\scriptstyle\pm1.73}$&  $62.17{\scriptstyle\pm2.48}$& \cellcolor{gray!15}$62.41{\scriptstyle\pm2.99}$&  $60.37{\scriptstyle\pm2.36}$&  $58.25{\scriptstyle\pm3.03}$&  $58.92{\scriptstyle\pm3.15}$
\\  \bottomrule
\vspace{-1.0cm}
\end{tabular}
}
\end{adjustbox}
\end{table*}

Table~\ref{tab:bn_ln} shows that both batch norm and layer norm can help with oversmoothing. Moreover, adding layer norm is in general better than adding batch norm. This is expected because the scaling effect caused by the propagation can be alleviated by normalizing across the node representations. Thus, the dispersion of the expected representations can be mitigated. On the other hand, batch norm normalizes across all the nodes, so it requires sacrificing the nodes that benefit (case 3) to compensate for the nodes that are prone to moving towards the other classes (case 1 \& case 2). As a result, batch norm is less effective in mitigating oversmoothing and leads to a bigger decrease in accuracy. 

Another finding is that both layer norm and batch norm lead to a significant accuracy decrease (2\%-3\%) on the heterophilous datasets. \texttt{+BN} has a clear accuracy drop even in the homophilous datasets. As Theorem~\ref{theorem: initial_stage} points out: higher heterophily level may result in sign flip. If the representations flip the sign, using batch norm or layer norm will not revert the sign, but they may instead encourage the representations to move towards the other class more.

Given the limitations shown above, we \textbf{do not} use either batch norm or layer norm in our proposed model, \method.

\subsection{More on the Initial \& Developing Stages}
\label{app:2stages}
Section \ref{exp:case study} shows how the node classification accuracy changes for nodes of different degrees with the number of layers on Citeseer. Here, we provide more details of this experiment and give the results on another dataset,  Cora.

\textbf{Datasets.} According to Thm.~\ref{theorem: initial_stage} and~\ref{theorem: developing stage}, in order to see both the initial and the developing stage, we need to use homophilous datasets. In  heterophilous datasets, most nodes satisfy case 1, so the initial stage does not exist. 

\textbf{Measurement of effective homophily $\pmb{\heterophilyil}$.} 
To estimate the effective homophily $\heterophilyil$, we measure the portion of neighbors that have the same ground truth label as $v_i$ and are correctly classified. 
Following our theory, we estimate $\heterophilyil$ \textbf{before} the last propagation takes place and analyze its impact on the accuracy of the final layer. 
In more detail, we obtain the node representations before the last propagation from a trained vanilla GCN~\cite{kipf2016semi} and then perform a linear transformation using the weight matrix and bias vector from the last layer. Then, we use the transformed representations to classify the neighbors of node $v_i$ and compute $\heterophilyil$. We note that we leverage the intermediate representations only for the estimation of $\heterophilyil$; the accuracy of the final layer is still measured using the outputs from the final layer. 

\setlength{\tabcolsep}{4pt}
\begin{wraptable}{r}{8.2cm}
    \centering
    \vspace{-0.3cm}
    \caption{Citeseer: Accuracy (Acc) and average effective homophily ($\Bar{\heterophilyil}$) for nodes with different degrees across various layers. Last layer of initial stage marked in gray.} 
    \vspace{0.1cm}
    \resizebox{!}{.38\linewidth}{
    \begin{tabular}{ccccccc} 
    \toprule
          & & \multicolumn{5}{c}{\bf Degrees} \\ \cline{3-7}
      \multicolumn{1}{c}{\bf  Layers} & 
      &   $[1,2]$           &   $[3,6]$            &   $[7,15]$   &   $[16,39]$ & $[40,99]$  \\
    \midrule
       \multirow{2}{*}{\bf 2}& Acc & $74.44$ & $78.51$ & $84.04$
        & $96.00$ & $100.00$  \\ & $\Bar{\heterophilyil}$ & $0.65$ & $0.67$ & $0.72$
        & $0.83$ & $0.91$  \\
        \midrule
       \multirow{2}{*}{\bf 3}& Acc & $70.92$ & $77.59$ & $83.03$
        & $92.75$ & $100.00$  \\& $\Bar{\heterophilyil}$  & $0.63$ & $0.66$ & $0.71$
        & $0.84$ & $0.92$  \\
        \midrule
       \multirow{2}{*}{\bf 4}& Acc & $60.54$ & $68.56$ & $77.63$
        & $94.00$ & $100.00$  \\& $\Bar{\heterophilyil}$  & \cellcolor{gray!15}$0.54$ & \cellcolor{gray!15}$0.58$ & \cellcolor{gray!15}$0.66$
        & \cellcolor{gray!15}$0.79$ & \cellcolor{gray!15}$0.84$  \\
        \midrule
       \multirow{2}{*}{\bf 5}& Acc & $41.66$ & $48.70$ & $56.97$
        & $61.33$ & $11.11$  \\& $\Bar{\heterophilyil}$  & $0.36$ & $0.38$ & $0.45$
        & $0.48$ & $0.11$  \\
        \midrule
       \multirow{2}{*}{\bf 6}& Acc & $23.61$ & $28.87$ & $41.17$
        & $31.83$ & $0.00$  \\& $\Bar{\heterophilyil}$  & $0.18$ & $0.19$ & $0.27$
        & $0.22$ & $0.00$  \\
	   \bottomrule
    \end{tabular}}
    \vspace{-0.7cm}
    \label{tab:case_study_citeseer}
\end{wraptable}

\textbf{Degree intervals.} To investigate the change in GCN accuracy with different layers for nodes with different degrees, we categorize the nodes in $n$ degree intervals. For the degree intervals, we use logarithmic binning (base 2). In detail, we denote the highest and lowest degree by $d_{max}$ and $d_{min}$, respectively, and let $\Omega \equiv \frac{\log_2{d_{max}}-\log_2{d_{min}}}{n}$. Then, we divide the nodes into $n$ intervals, where the $j$-th interval is defined as: $[d_{min}\cdot 2^{(j-1)\Omega}, d_{min}\cdot 2^{j\Omega})$.


\textbf{Dataset: Citeseer.}
Figure~\ref{fig:phase_transition} and Table~\ref{tab:case_study_citeseer} show how the accuracy changes with the number of layers for different node degree intervals. We observe that in the initial stage, the accuracy increases as the degree and $\Bar{\heterophilyil}$ increase. However, in the developing stage, the accuracy of high-degree nodes drops more sharply than that of low-degree nodes. 

\textbf{Dataset: Cora.}  The results for Cora are shown in 
Figure~\ref{fig:case_study_cora} and Table~\ref{tab:case_study_cora}.  
In the initial stage, the nodes with lower degrees usually have lower accuracy. One exception is the nodes with degrees in the 
range $[3, 7]$. These nodes have higher accuracy because the average effective homophily $\Bar{\heterophilyil}$ of that degree group is the second highest. In the developing stage, the accuracy of the high-degree nodes drops more than the accuracy of the remaining node groups. 

GCN's behavior on both Citesser and Cora datasets verifies our conjecture based on our theorems in \S~\ref{subsec:node_profile}.

\begin{minipage}{\textwidth}
\begin{minipage}[c]{0.55\textwidth}
\centering
\includegraphics[scale=0.18, bb=800 400 450 200]{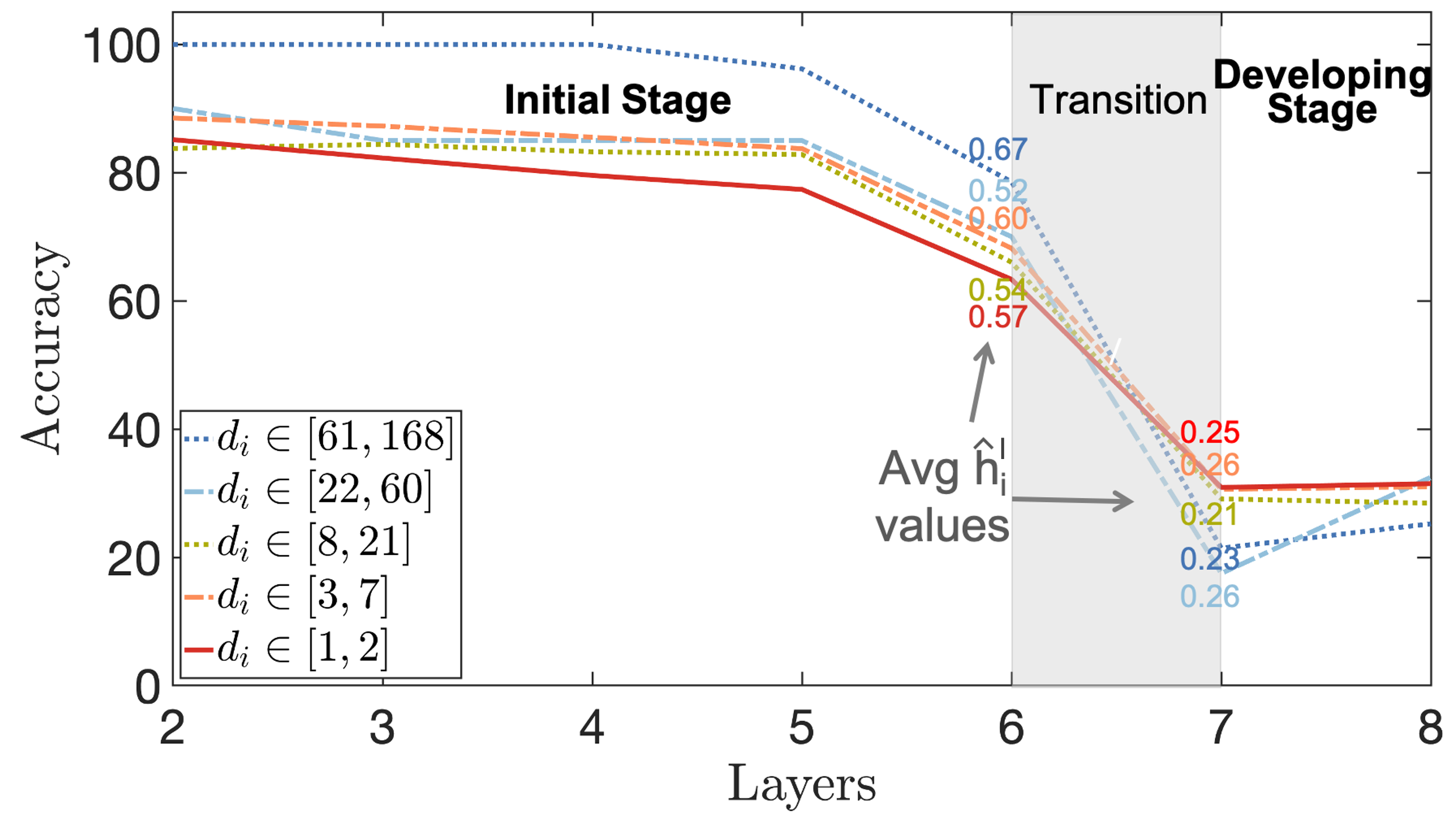}
\captionsetup{type=figure}
\vspace{2.5cm}
\captionof{figure}{Cora: Accuracy per (logarithmic) degree bin.}
\label{fig:case_study_cora}
\end{minipage}
\begin{minipage}[c]{0.45\textwidth}
\setlength{\tabcolsep}{5pt}
    \centering
    \captionsetup{type=table}
    \captionof{table}{Cora: Accuracy and average effective homophily ($\Bar{\heterophilyil}$) for nodes with different degrees across different layers. Last layer of initial stage marked in gray.}
    \resizebox{!}{.45\linewidth}{
    \begin{tabular}{ccccccc} 
    \toprule
      & & \multicolumn{5}{c}{\bf Degrees} \\ \cline{3-7}
      \multicolumn{1}{c}{\bf Layers} & 
      &   $[1,2]$           &   $[3,7]$            &   $[8,21]$   &   $[22,60]$ & $[61,168]$  \\
    \midrule
       \multirow{2}{*}{\bf 2}& Acc & $85.14$ & $88.52$ & $83.78$
        & $90.00$ & $100.00$  \\ & $\Bar{\heterophilyil}$ & $0.77$ & $0.77$ & $0.72$
        & $0.63$ & $0.81$  \\
        \midrule
       \multirow{2}{*}{\bf 3}& Acc & $82.28$ & $87.26$ & $84.42$
        & $85.00$ & $100.00$  \\& $\Bar{\heterophilyil}$  & $0.77$ & $0.78$ & $0.73$
        & $0.64$ & $0.84$  \\
        \midrule
       \multirow{2}{*}{\bf 4}& Acc & $79.57$ & $85.52$ & $83.26$
        & $85.00$ & $100.00$  \\& $\Bar{\heterophilyil}$  & $0.75$ & $0.76$ & $0.71$
        & $0.63$ & $0.80$  \\
        \midrule
        \multirow{2}{*}{\bf 5}& Acc & $77.38$ & $83.75$ & $82.83$
        & $85.00$ & $96.19$  \\& $\Bar{\heterophilyil}$  & $0.72$ & $0.73$ & $0.68$
        & $0.56$ & $0.83$  \\
        \midrule
        \multirow{2}{*}{\bf 6}& Acc & $63.35$ & $68.20$ & $66.01$
        & $70.00$ & $78.57$  \\& $\Bar{\heterophilyil}$  & \cellcolor{gray!15}$0.57$ & \cellcolor{gray!15}$0.60$ & \cellcolor{gray!15}$0.54$
        & \cellcolor{gray!15}$0.52$ & \cellcolor{gray!15}$0.67$  \\
        \midrule
       \multirow{2}{*}{\bf 7}& Acc & $30.91$ & $30.62$ & $29.12$
        & $17.50$ & $21.43$  \\& $\Bar{\heterophilyil}$  & $0.25$ & $0.26$ & $0.21$
        & $0.26$ & $0.23$  \\
        \midrule
       \multirow{2}{*}{\bf 8}& Acc & $31.48$ & $31.02$ & $28.44$
        & $32.50$ & $25.24$  \\& $\Bar{\heterophilyil}$  & $0.24$ & $0.26$ & $0.20$
        & $0.28$ & $0.24$  \\
	   \bottomrule
    \end{tabular}
        \vspace{-1cm}
    }
\label{tab:case_study_cora}
\end{minipage}
    \vspace{-0.1cm}
\end{minipage}

{
\subsection{Complexity analysis}
We first analyze the time complexity of the forward path \method. For degree correction, we need to compute the $r_{ij}$. However, we do not need to compute all of them, and we only need to compute $r_{ij}$ for node pairs who are linked. The time complexity is $O(|\gE|)$, but it is a one-time computation, the results of which can be saved. We compute $\tau_{ij}^{l}$ based on the learned weights $\lambda_0^l, \lambda_1^{l}$. The time complexity is $O(|\gE| \cdot L)$, where $L$ is the number of layers. The time complexity to compute the signed function is $O(|\gE| \cdot H \cdot L)$, where $H$ is the hidden dimension of representations, and we only compute the cosine similarity between the nodes that are linked. We also need to compute the multiplication of the propagation matrix and representation matrices. The time complexity is $O(|\gV|^2 \cdot H \cdot L)$, . Similarly, the time complexity of the multiplication of representation matrices and the weight matrices is $O(|\gV|\cdot H^2 \cdot L)$. The total time complexity would be $O(\text{max}(|\gV|^2, |\gE|)\cdot H \cdot L)$.
The complexity of \method resembles the complexity of attention-based model. Thus we compare with GAT~\citep{velivckovic2017graph} the total time to train and test on 4 larger homophilous and heterophilous datasets. We run the training and testing for 10 times.
}

\setlength{\tabcolsep}{4pt}
\begin{table*}[h]
    \centering
    \vspace{-0.2cm}
    \caption{
    Training and test time (m:minutes, s:seconds) for 10 runs. Shorter time is colored in grey.
    }
    \label{tab:scalability}
    \vspace{-0.15cm}
    \begin{adjustbox}{width=.5\textwidth}{\footnotesize
    \begin{tabular}{p{2cm} p{2cm} p{2cm} p{2cm} p{2cm}} 
    \toprule
       &     \multicolumn{1}{c}{\texttt{\bf Actor}}            &      \multicolumn{1}{c}{\texttt{\bf Chameleon}} &   \multicolumn{1}{c}{\texttt{\bf Citeseer}}                       &   \multicolumn{1}{c}{\texttt{\bf Cora}} \\
		\multicolumn{1}{c}{\textbf{\#Nodes}} &  \multicolumn{1}{c}{7,600} &  \multicolumn{1}{c}{2,277} &  \multicolumn{1}{c}{3,327}  & \multicolumn{1}{c}{2,708} \\
		\multicolumn{1}{c}{\textbf{\#Edges}} &  \multicolumn{1}{c}{26,752} &  \multicolumn{1}{c}{31,421} &  \multicolumn{1}{c}{4,676}  & \multicolumn{1}{c}{5,278} \\
    \midrule
       \multicolumn{1}{c}{\method} & \multicolumn{1}{c}{\cellcolor{gray!15}$7m24s$} & \multicolumn{1}{c}{\cellcolor{gray!15}$2m29s$} & \multicolumn{1}{c}{\cellcolor{gray!15}
       $2m9s$} & \multicolumn{1}{c}{$5m47s$}  \\
       \multicolumn{1}{c}{GAT} & \multicolumn{1}{c}{$8m14s$} & \multicolumn{1}{c}{$7m34s$} & \multicolumn{1}{c}{$9m39s$} & \multicolumn{1}{c}{\cellcolor{gray!15}$4m11s$}  \\
       
	   \bottomrule
    \end{tabular}}
    \end{adjustbox}
    \vspace{-0.3cm}
\end{table*}
In general, \method runs faster than GAT because degree correction and signed messages in \method learn fewer parameters.
\section{Hyperparameters and Parameters}
\label{app:hyperparameter}
\subsection{Hyperparameter settings}
\textbf{Experiments for Table~\ref{tab:real-results} \& Table~\ref{tab:oversmoothing}}

For the baselines, we set the same hyperparameters that are provided by the original papers or the authors’ github repositories, and we match the results they reported in their respective papers. In our experiments, we find that the original hyperparameters set by the authors are already well-tuned. 

All the models use Adam as the optimizer. GAT sets the initial learning rate as 0.005 and Geom-GCN uses a custom learning scheduler. All the other models (include \method) use the initial learning rate 0.01.

For \method, we use the following hyperparameters:
\begin{itemize}
    \item $k$ in the decaying aggregation: 3
\end{itemize}

We tune the parameters in the following ranges:
\begin{itemize}
    \item Dropout rate:  [0.0, 0.7]
    \item Weight decay:  [1e-7, 1e-2]
    \item Hidden units: \{8, 16, 32, 64, 80\}
    \item Decay rate $\eta$: [0.0, 1.5]
\end{itemize}

\textbf{Experiments for Table~\ref{tab:ablation} and Table~\ref{tab:bn_ln}}

The hyperparameters that are used in all the models (Base, \texttt{+deg}, \texttt{+sign}, \texttt{+deg,sign}, \texttt{+BN}, \texttt{+LN}) are set to be the same and they are tuned for every dataset. Those common hyperparameters are:
\begin{itemize}
    \item Dropout rate: [0.0, 0.7]
    \item Weight decay: [1e-7, 1e-2]
    \item Hidden units: \{8, 16, 32, 64, 80\}
\end{itemize}
\subsection{Initialization of parameters}
\textbf{Initialization}

For \method, we adopt the following parameter initialization in the experiments for Table~\ref{tab:real-results} \& Table~\ref{tab:oversmoothing}
\begin{itemize}
    \item Initialization of $\lambda_0^l$ and $\lambda_1^l$: 0.5 and 0, respectively
    \item Initialization of $\alpha^l$, $\beta_0^l$, $\beta_1^l$ and $\beta_2^l$: 2, 0, 0, 0, respectively.
\end{itemize}
We initialize 
{\small $\beta_{\{0,1,2\}}^l=0$} 
in Eq.\ (6) because, after applying softmax, 
{\small $\hat{\beta}^l_{\{0,1,2\}}=1/3$}; 
this ensures \ul{equal contributions} from positive and negative neighbors and themselves (and sum=1).
We initialize $\lambda_1^l=0$ in Eq.\ (9) following the common practice for initializing the bias. 
We set $\alpha^l=2$ and $\lambda_0^l=0.5$ in Eq.\ (6) and Eq.\ (9), because when $r_{ij} \to +\infty$, the degree correction (including global scaling) is: {\small $\hat{\alpha^l}\tau_{ij}^l=\text{softplus}(2) \cdot \text{softplus}(0.5 \cdot (0-1) + 0) \approx 1$}. As we mention in Thm. 1, when the homophily level is high, nodes with large $\overline{r_i}$ may benefit (case 3), thus we \ul{do not want to compensate for these nodes and would like to keep {\small $\hat{\alpha^l}\tau_{ij}^l$} close to 1}. 
\subsection{Parameters after training}
\begin{wrapfigure}{r}{0.5\textwidth}
  \begin{center}
    \includegraphics[scale=0.06,bb=3000 1000 380 180]{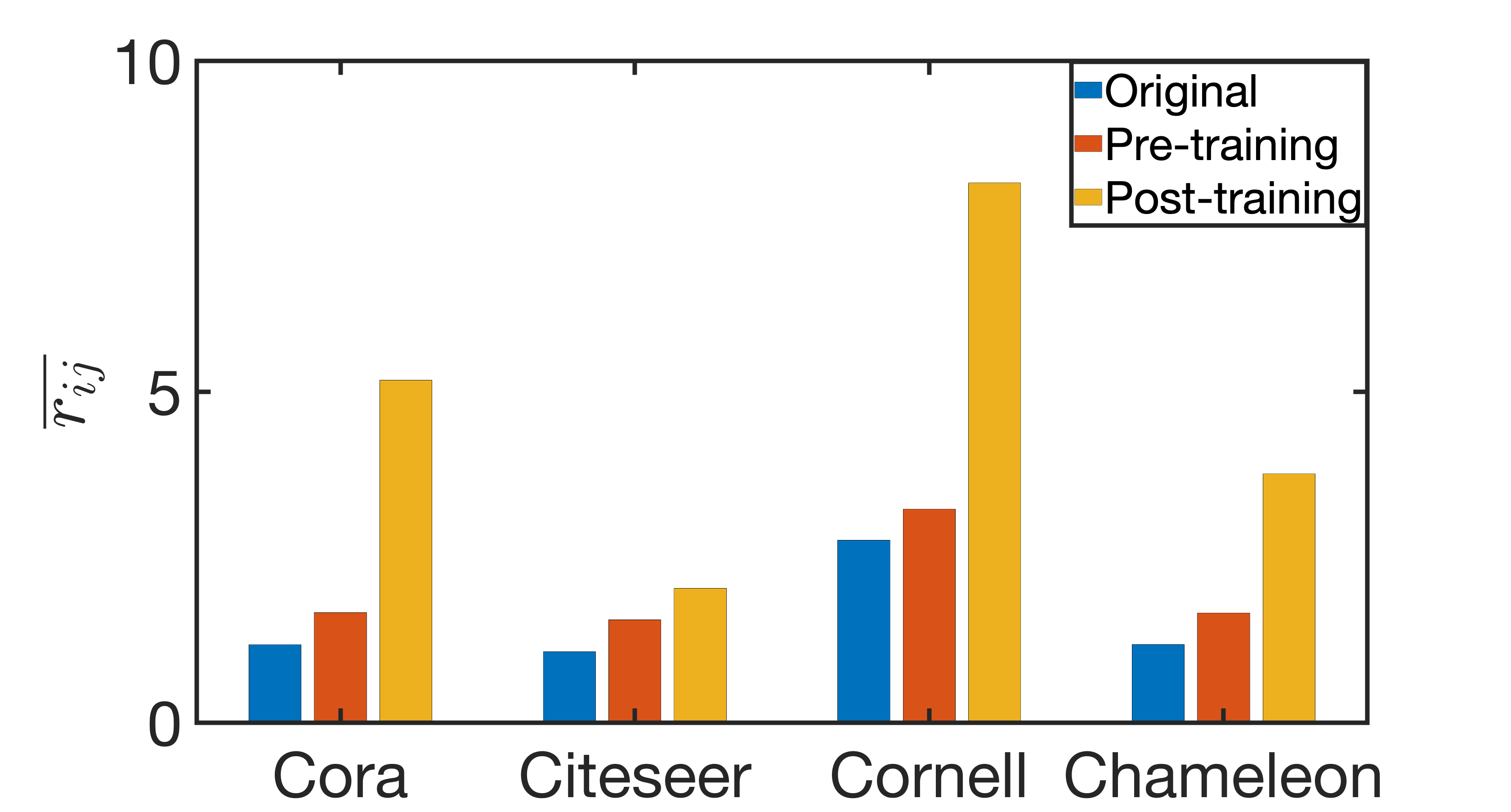}
  \end{center}
  \vspace{1.8cm}
  \caption{Original $\overline{r_{ij}}$ (avg. $r_{ij}$) 
  and corrected (pre- and post-training).} 
  \label{fig:parameters}
\end{wrapfigure}
Figure~\ref{fig:parameters} shows the original and corrected $\overline{r_{ij}}$ (avg $r_{ij}$). Corrected $r_{ij}$ are given by {\small $\hat{\alpha^l}\tau_{ij}^lr_{ij}$}, which are a combination of global scaling and local degree correction. Because {\small $\beta_{\{0,1,2\}}^l$} sum to 1 and do not change global scaling, they are not considered in the corrected $r_{ij}$.
As can be seen in Fig.~\ref{fig:parameters}, after the training, \ul{GGCN learns to increase $r_{ij}$}, which satisfies our theorems.

\section{Table 2 with larger fonts}
\label{app:table 2}
\begin{sidewaystable}[h]
\centering
\caption{Model performance for different layers: mean accuracy $\pm$ stdev over different data splits. Per dataset and GNN model, we also report the layer at which the best performance (given in Table~\ref{tab:real-results}) is achieved. `OOM': out of memory; `INS': numerical instability. }
\label{app-tab:table 2}
\begin{adjustbox}{width=1\linewidth}
{\footnotesize
\begin{tabular}{r c ccccccccccccccc}
\toprule
                                      \textbf{Layers} 
                                      && \textbf{2}           & \multicolumn{1}{c}{\textbf{4}} & \multicolumn{1}{c}{\textbf{8}} & \multicolumn{1}{c}{\textbf{16}} & \multicolumn{1}{c}{\textbf{32}} & \multicolumn{1}{c}{\textbf{64}} &
                                      \multicolumn{1}{c}{\textbf{Best}}
                                      &&  \multicolumn{1}{c}{\textbf{2}}           & \multicolumn{1}{c}{\textbf{4}} & \multicolumn{1}{c}{\textbf{8}} & \multicolumn{1}{c}{\textbf{16}} & \multicolumn{1}{c}{\textbf{32}} & \multicolumn{1}{c}{\textbf{64}}  &
                                      \multicolumn{1}{c}{\textbf{Best}} \\ \cmidrule{1-1} \cmidrule{3-9} \cmidrule{11-17} 
&& \multicolumn{7}{c}{\textbf{Cora} ($h$=0.81)}     & & \multicolumn{7}{c}{\textbf{Citeseer} ($h$=0.74)}                                                                         \\ 
\cline{3-9} \cline{11-17}
                                    
                                        \multicolumn{1}{l}{GGCN (\textbf{ours})}    &&   
                                        $87.00{\scriptstyle\pm1.15}$         & {$ 87.48{\scriptstyle\pm1.32}$}           & {$ 87.63{\scriptstyle\pm1.33}$}           & {$ 87.51{\scriptstyle\pm1.19}$}            & 
                                        {$ 87.95{\scriptstyle\pm1.05}$}            & {$ 87.28{\scriptstyle\pm1.41}$}            & 
                                        32
                                           & &  
                                        $76.83{\scriptstyle\pm1.82}$ & $76.77{\scriptstyle\pm1.48}$ & $76.91{\scriptstyle\pm1.56}$ & $76.88{\scriptstyle\pm1.56}$& $76.97{\scriptstyle\pm1.52}$ & $76.65{\scriptstyle\pm1.38}$ & 
                                        10 \\ 
                                        
                            \multicolumn{1}{l}{GPRGNN}    &&  
                                      \multicolumn{1}{c}{$87.93{\scriptstyle\pm1.11}$}      &        $87.95{\scriptstyle\pm1.18}$           & \multicolumn{1}{c}{$87.87{\scriptstyle\pm1.41}$}           & \multicolumn{1}{c}{$87.26{\scriptstyle\pm1.51}$}            & \multicolumn{1}{c}{$87.18{\scriptstyle\pm1.29}$}            & \multicolumn{1}{c}{$87.32{\scriptstyle\pm1.21}$} & 4 &   &                   
                                      $77.13{\scriptstyle\pm1.67}$  & $77.05{\scriptstyle\pm1.43}$&  $77.09{\scriptstyle\pm1.62}$&  $76.00{\scriptstyle\pm1.64}$&  $74.97{\scriptstyle\pm1.47}$&  $74.41{\scriptstyle\pm1.65}$
                            & 2
                            \\
                            \multicolumn{1}{l}{H2GCN*} &&  
                                      $87.87{\scriptstyle\pm1.20}$  &   $86.10{\scriptstyle\pm1.51}$&   $86.18{\scriptstyle\pm2.10}$&    OOM                    &       OOM                     &      OOM                      &     2
                                      & &    
                                    $76.90{\scriptstyle\pm1.80}$  & 
                                      $76.09{\scriptstyle\pm1.54}$  & 
                                      $74.10{\scriptstyle\pm1.83}$  &  
                                      OOM     &  
                                      OOM     &  
                                      OOM     &     1
                                      
                        \\      
                                     \multicolumn{1}{l}{GCNII*}    &&  
                                      $85.35{\scriptstyle\pm1.56}$                & {$85.35{\scriptstyle\pm1.48}$}           & {$86.38{\scriptstyle\pm0.98}$}           & {$87.12{\scriptstyle\pm1.11}$}            & {$87.95{\scriptstyle\pm1.23}$}            & {$88.37{\scriptstyle\pm1.25}$}      &       
                                      64  & &
                                      $75.42{\scriptstyle\pm1.78}$  & $75.29{\scriptstyle\pm1.90}$&  $76.00{\scriptstyle\pm1.66}$&  $76.96{\scriptstyle\pm1.38}$&  $77.33{\scriptstyle\pm1.48}$&  $77.18{\scriptstyle\pm1.47}$ & 32
                                      \\  
                                      
                        \multicolumn{1}{l}{PairNorm}    &&   
                                        $ 85.79{\scriptstyle\pm1.01}$         & \multicolumn{1}{c}{$ 85.07{\scriptstyle\pm0.91}$}    &        \multicolumn{1}{c}{$ 84.65{\scriptstyle\pm1.09}$}     &       \multicolumn{1}{c}{$ 82.21{\scriptstyle\pm2.84}$}      &     \multicolumn{1}{c}{$ 60.32{\scriptstyle\pm8.28}$}       &    \multicolumn{1}{c}{$ 44.39{\scriptstyle\pm5.60}$}   & 2 & & 
                                        $73.59{\scriptstyle\pm1.47}$ & $72.62{\scriptstyle\pm1.97}$ & $72.32{\scriptstyle\pm1.58}$ & $59.71{\scriptstyle\pm15.97}$& $27.21{\scriptstyle\pm10.95}$ & $23.82{\scriptstyle\pm6.64}$ & 2                                        \\               
                        \multicolumn{1}{l}{Geom-GCN*} &&
                          $85.35{\scriptstyle\pm1.57}$&   $21.01{\scriptstyle\pm2.61}$&   $13.98{\scriptstyle\pm1.48}$&   $13.98{\scriptstyle\pm1.48}$&   $13.98{\scriptstyle\pm1.48}$&   $13.98{\scriptstyle\pm1.48}$&   
                          2
                         &  & 
                         $78.02{\scriptstyle\pm1.15}$&  $23.01{\scriptstyle\pm1.95}$&   $7.23{\scriptstyle\pm0.87}$&   $7.23{\scriptstyle\pm0.87}$&   $7.23{\scriptstyle\pm0.87}$&   $7.23{\scriptstyle\pm0.87}$
                         &   2 
                         \\ 
                        \multicolumn{1}{l}{GCN} &&
                        $86.98{\scriptstyle\pm1.27}$ & $83.24{\scriptstyle\pm1.56}$ & $31.03{\scriptstyle\pm3.08}$ &
                        $31.05{\scriptstyle\pm2.36}$ & $30.76{\scriptstyle\pm3.43}$ & $31.89{\scriptstyle\pm2.08}$ & 2
                         &  & 
                        $76.50{\scriptstyle\pm1.36}$ & $64.33{\scriptstyle\pm8.27}$ & $24.18{\scriptstyle\pm1.71}$ &
                               $23.07{\scriptstyle\pm2.95}$ & $25.3{\scriptstyle\pm1.77}$ & $24.73{\scriptstyle\pm1.66}$ & 2  
                         \\              
                        
                         \multicolumn{1}{l}{GAT} &&
                         $87.30{\scriptstyle\pm1.10}$ & $86.50{\scriptstyle\pm1.20}$ & $84.97{\scriptstyle\pm1.24}$ &
                              INS  & INS & INS & 2
                         &  & 
                        $76.55{\scriptstyle\pm1.23}$ &$75.33{\scriptstyle\pm1.39}$&    $66.57{\scriptstyle\pm5.08}$         &    INS                             &    INS                             &       INS                          & 2  
                         \\
                         
                         \midrule 
&& \multicolumn{7}{c}{\textbf{Cornell ($h$=0.3)}} && \multicolumn{7}{c}{\textbf{Chameleon ($h$=0.23)}}                                                                         \\ 
\cline{3-9} \cline{11-17}
                                    
                        \multicolumn{1}{l}{GGCN (\textbf{ours})}    &&   
                        $83.78{\scriptstyle\pm6.73}$ & $83.78{\scriptstyle\pm6.16}$ & $84.86{\scriptstyle\pm5.69}$ &
                        $83.78{\scriptstyle\pm6.73}$ & 
                        $83.78{\scriptstyle\pm6.51}$ & 
                        $84.32{\scriptstyle\pm5.90}$ & 
                               6
                        & &  
                        $70.77{\scriptstyle\pm1.42}$ & $69.58{\scriptstyle\pm2.68}$ & $70.33{\scriptstyle\pm1.70}$ & $70.44{\scriptstyle\pm1.82}$& $70.29{\scriptstyle\pm1.62}$ & $70.20{\scriptstyle\pm1.95}$ & 
                        5                
                        \\ 
            \multicolumn{1}{l}{GPRGNN}    && $76.76{\scriptstyle\pm8.22}$ & $77.57{\scriptstyle\pm7.46}$ & $80.27{\scriptstyle\pm8.11}$ &
                              $78.38{\scriptstyle\pm6.04}$ & $74.59{\scriptstyle\pm7.66}$ & $70.00{\scriptstyle\pm5.73}$    & 8 & &
$46.58{\scriptstyle\pm1.771}$ & $45.72{\scriptstyle\pm3.45}$ & $41.16{\scriptstyle\pm5.79}$ & $39.58{\scriptstyle\pm7.85}$& $35.42{\scriptstyle\pm8.52}$ & $36.38{\scriptstyle\pm2.40}$ & 2
            \\
            \multicolumn{1}{l}{H2GCN*} &&  
                        $81.89{\scriptstyle\pm5.98}$ & $82.70{\scriptstyle\pm6.27}$ & $80.27{\scriptstyle\pm6.63}$ &
                               OOM & OOM & OOM & 1  
                        & &    
                        $59.06{\scriptstyle\pm1.85}$ & $60.11{\scriptstyle\pm2.15}$ & OOM &                          OOM &                          OOM &                          OOM &                          4            
                        \\  
            
                        \multicolumn{1}{l}{GCNII*} && 
                        $67.57{\scriptstyle\pm11.34}$ & $64.59{\scriptstyle\pm9.63}$ & $73.24{\scriptstyle\pm5.91}$ &
                               $77.84{\scriptstyle\pm3.97}$ & $75.41{\scriptstyle\pm5.47}$ & $73.78{\scriptstyle\pm4.37}$ & 16   
                        & &
                        $61.07{\scriptstyle\pm4.10}$ & $63.86{\scriptstyle\pm3.04}$ & $62.89{\scriptstyle\pm1.18}$ & $60.20{\scriptstyle\pm2.10}$& $56.97{\scriptstyle\pm1.81}$ & $55.99{\scriptstyle\pm2.27}$ & 4  
                        \\ 
                        \multicolumn{1}{l}{PairNorm}    && $50.27{\scriptstyle\pm7.17}$ & $53.51{\scriptstyle\pm8.00}$ & $58.38{\scriptstyle\pm5.01}$ &
                              $58.38{\scriptstyle\pm3.01}$ & $58.92{\scriptstyle\pm3.15}$ & $58.92{\scriptstyle\pm3.15}$          & 32 & &
$62.74{\scriptstyle\pm2.82}$ & $59.01{\scriptstyle\pm2.80}$ & $54.12{\scriptstyle\pm2.24}$ & $46.38{\scriptstyle\pm2.23}$& $46.78{\scriptstyle\pm2.26}$ & $46.27{\scriptstyle\pm3.24}$ & 2
\\  
                        
                        \multicolumn{1}{l}{Geom-GCN*} &&
                        $60.54{\scriptstyle\pm3.67}$&  $23.78{\scriptstyle\pm11.64}$& $12.97{\scriptstyle\pm2.91}$&  $12.97{\scriptstyle\pm2.91}$&  $12.97{\scriptstyle\pm2.91}$&  $12.97{\scriptstyle\pm2.91}$&  
                        2  
                         &  & 
                        $60.00{\scriptstyle\pm2.81}$&  $19.17{\scriptstyle\pm1.66}$& $19.58{\scriptstyle\pm1.73}$&  $19.58{\scriptstyle\pm1.73}$&  $19.58{\scriptstyle\pm1.73}$&  $19.58{\scriptstyle\pm1.73}$
                        &  2 
                        \\
                         \multicolumn{1}{l}{GCN} &&
                        $60.54{\scriptstyle\pm5.30}$ & $59.19{\scriptstyle\pm3.30}$ & $58.92{\scriptstyle\pm3.15}$ &
                        $58.92{\scriptstyle\pm3.15}$ & 
                        $58.92{\scriptstyle\pm3.15}$ & 
                        $58.92{\scriptstyle\pm3.15}$ & 2
                               &  & 
                        $64.82{\scriptstyle\pm2.24}$ & $53.11{\scriptstyle\pm4.44}$ & $35.15{\scriptstyle\pm3.14}$ &
                               $35.39{\scriptstyle\pm3.23}$ & $35.20{\scriptstyle\pm3.25}$ & $35.50{\scriptstyle\pm3.08}$ & 2 
                         \\ 
                         \multicolumn{1}{l}{GAT} &&
                        $61.89{\scriptstyle\pm5.05}$ &      $58.38{\scriptstyle\pm4.05}$ &     $58.38{\scriptstyle\pm3.86}$                           
                        &   INS                    
                        &   INS                         
                        &   INS &  
                        2 
                         &  & 
                        $60.26{\scriptstyle\pm2.50}$ &   $48.71{\scriptstyle\pm2.96}$                             &  $35.09{\scriptstyle\pm3.55}$                              &      INS                           &    INS                             &   INS                              &  2 
                         
                        \\ \bottomrule
\end{tabular}
}
\end{adjustbox}
\end{sidewaystable}


\end{document}